%% file: IEEE-conference-template-062824.tex
\documentclass[conference]{IEEEtran}
\IEEEoverridecommandlockouts

\usepackage{microtype}
\usepackage{graphicx}
\usepackage{subfigure}
\usepackage{booktabs} 

\usepackage{hyperref}%
\usepackage{url}
\usepackage{amsmath}
\usepackage{algorithm}
\usepackage{algorithmic}
\usepackage{graphicx}
\usepackage{wrapfig}
\usepackage{booktabs}
\usepackage{enumitem}
\usepackage{multirow}
\usepackage{amsthm}
\usepackage{dsfont}
\usepackage[capitalize,noabbrev]{cleveref}
\usepackage[numbers]{natbib}
\usepackage{color}

\input{macro.tex}

\newcommand{\problemname}{deep unlearning}
\newcommand{\problemnamepassive}{deeply unlearnt}
\newcommand{\problemnameactive}{deeply unlearn}
\newcommand{\problemnamethird}{deeply unlearns}
\newcommand{\problemnameing}{deeply unlearning}
\newcommand{\problemnamecap}{Deep unlearning}
\newcommand{\dataname}{Eval-DU}

\def\BibTeX{{\rm B\kern-.05em{\sc i\kern-.025em b}\kern-.08em
    T\kern-.1667em\lower.7ex\hbox{E}\kern-.125emX}}
\begin{document}

\title{Evaluating Deep Unlearning in Large Language Models}

\author{\IEEEauthorblockN{Ruihan Wu}
\IEEEauthorblockA{UC San Diego\\
\textit{\texttt{ruw076@ucsd.edu}}
}
\and
\IEEEauthorblockN{Chhavi Yadav}
\IEEEauthorblockA{UC, San Diego}
\textit{\texttt{cyadav@ucsd.edu}}
\and
\IEEEauthorblockN{Russ Salakhutdinov}
\IEEEauthorblockA{CMU}
\textit{\texttt{rsalakhu@cs.cmu.edu}}
\and
\IEEEauthorblockN{Kamalika Chaudhuri}
\IEEEauthorblockA{UC, San Diego
}
\textit{\texttt{kamalika@ucsd.edu}}
}


\maketitle

\begin{abstract}
Machine unlearning has emerged as an important component in developing safe and trustworthy models. 
Prior work on fact unlearning in LLMs has mostly focused on removing a specified target fact robustly, but often overlooks its deductive connections to other knowledge.
We propose a new setting for fact unlearning, \textit{\problemname}, where the goal is not only to remove a target fact but also to prevent it from being deduced via retained knowledge in the LLM and logical reasoning.
We propose three novel metrics: \textit{Success-DU} and \textit{Recall} to measure unlearning efficacy, and \textit{Accuracy} to measure the remainder model utility.
To benchmark this setting, we leverage both (1) an existing real-world knowledge dataset, MQuAKE, that provides one-step deduction instances, and (2) newly construct a novel semi-synthetic dataset, Eval-DU, that allows multiple steps of realistic deductions among synthetic facts.
Experiments reveal that current methods struggle with deep unlearning: they either fail to deeply unlearn, or excessively remove \textit{unrelated} facts.
Our results suggest that targeted algorithms may have to be developed for robust/deep fact unlearning in LLMs.
\end{abstract}


\input{sections/intro_rebuttal}
\input{sections/preliminary}
\input{sections/formulation}
\input{sections/synthetic_dataset}
\input{sections/experiment}
\input{sections/related_work}
\input{sections/discussion}

\section*{LLM usage considerations}
We have two LLM usages in this paper: use LLM to generate the texts in our synthetic dataset and use LLMs to refine the grammar and clarity of our writing. The core ideas and research progress are developed independently through our own study and investigation.


\bibliography{reference}
\bibliographystyle{plainnat}

\newpage
\newpage
\appendix

\input{sections/appendix}

\end{document}

%% file: macro.tex
\newcommand{\calR}{\mathcal{R}}
\newcommand{\calO}{\mathcal{O}}
\newcommand{\calA}{\mathcal{A}}

\newcommand{\calK}{\mathcal{K}}
\newcommand{\calM}{\mathcal{M}}

\newcommand{\calI}{\mathcal{I}}

\newcommand{\calT}{\mathcal{T}}

\newtheorem{theorem}{Theorem}

\newtheorem{definition}{Definition}

%% file: sections/intro_rebuttal.tex
\section{Introduction}
Large language models (LLMs) of today are trained on massive amounts of uncurated data obtained from the internet. 
Machine unlearning~\citep{liu2024rethinking} in LLMs aims to remove specific pieces of data, concepts, or facts from these models in an efficient way rather than retraining the model from scratch.
The diverse definitions of unlearning (data, concept or fact unlearning) are tailored to different use cases.
For instance, compliance with regulations such as the GDPR~\citep{gdpr} mandates the removal of a user's record~\citep{ginart2019making, guo2020certified}. 
Similarly, unlearning can be used to address concerns that models retain copyrighted material \citep{eldan2023s, dou2024avoiding} or offensive content~\citep{yao2023large}.

In this paper, we consider the problem of unlearning \textit{facts} from an LLM, which is important in scenarios with privacy requirements.
Research has shown that LLMs can memorize personal and sensitive information~\citep{carlini2021extracting, nasr2023scalable}, including relationships, work histories, and personal addresses. 
Such information can be readily accessed by LLM users, posing significant privacy risks and raising ethical concerns over uncontrolled exposure of private data. This motivates the need to unlearn facts -- after unlearning, the target fact cannot be reconstructed.

\begin{figure}[!t]
\centering
\includegraphics[width=\linewidth]{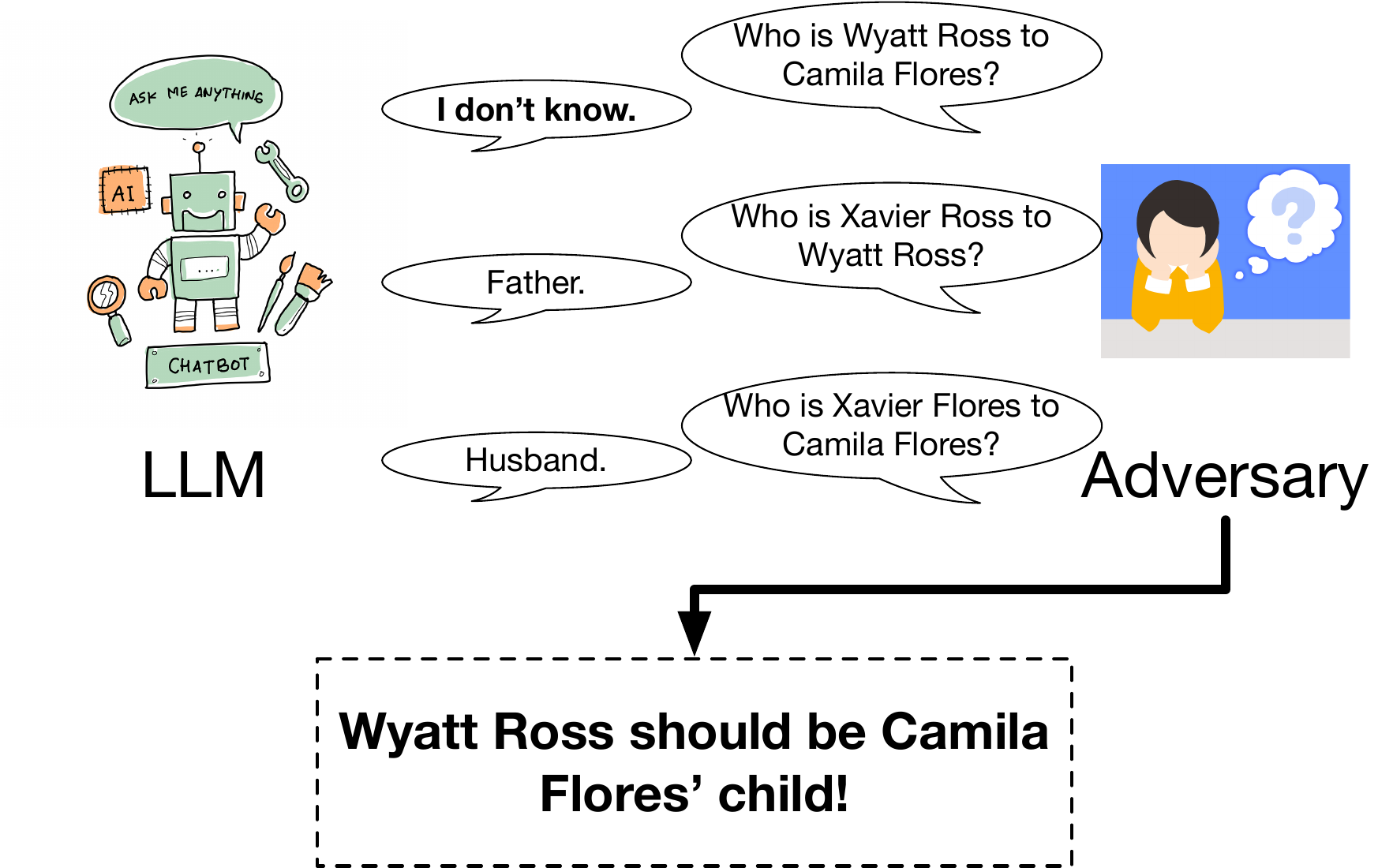}	
\caption{An example that unlearning only the target fact is insufficient. The successful extraction of \textit{``Wyatt Ross's father is Xavier Ross"} and \textit{``Camila Flores's husband is Xavier Ross"} can imply the target fact.}
\label{fig:intro}
\end{figure}

Some prior works~\citep{patil2024can,tofu2024,wang2024large} have looked at the problem of fact unlearning, but the focus has been on removing the specified target fact \textit{itself} in a robust way. 
However, this can be superficial -- LLMs not only know single facts in isolation, but many connected facts -- and the fact that has been unlearnt likely can be deduced from retained facts in the model. Thus, successful unlearning in this setting should also remove other facts that imply the fact to be unlearnt. As a concrete example, consider Figure~\ref{fig:intro}. Here, the target fact \textit{``Camila Flores's child is Wyatt Ross"} can be deduced from fact A \textit{``Wyatt Ross's father is Xavier Ross"} and fact B \textit{``Camila Flores's husband is Xavier Ross"}.
If the LLM only unlearns the target fact but retains A and B, 
this is insufficient as an adversary who extracts A and B  from the LLM can deduce the target fact.

We consider a new setting for unlearning, which we call \textit{\problemname}, and investigate to what extent current unlearning methods succeed in this setting. 
We formulate a set of facts as the knowledge base consisting of the triplet facts, and consider the structured logical rules as the deducing process when the adversary tries to reconstruct the target fact from the retained facts.
Then we define deep unlearning: the fact is \problemnamepassive\ if the target fact cannot be deduced from the retained facts in the LLM through the given logical rules.
We further propose three metrics, Success-DU, Recall and Accuracy: Success-DU and Recall for evaluating the efficacy of deep unlearning, and Accuracy for measuring to what extent other irrelevant facts are retained by the unlearning process.

To enable systematic evaluation of deep unlearning, we establish two complementary benchmarks that offers different aspects.
The first is based on the MQuAKE dataset~\citep{zhong2023mquake}.
It supports the evaluation in a realistic setting, where deep unlearning is assessed through single-step deductions over real-world facts encoded in pretrained models.
To further evaluate deep unlearning in more challenging scenarios involving multi-step deductions and under fully controlled conditions with complete knowledge base observability, we introduce a new \emph{semi-synthetic} dataset, Eval-DU.
Eval-DU is built on \emph{synthetic} family relationships and biographical facts as well as \emph{realistic} logical rules between family relationships.
The richer rule set induces more complex and longer deductive chains, which creates a more challenging testbed for deep unlearning.
The full observability of the synthetic knowledge space enables reliable evaluation.
Together, MQuAKE and Eval-DU form a systematic and robust testbed for studying deep unlearning in both realistic and fully controlled environments.

We use our benchmarks to evaluate four representative unlearning methods: Gradient Ascent, Negative Preference Optimization, Task Vector, and Who's Harry Potter.
The evaluation is on the MQuAKE dataset with two LLMs (GPT-J-6B and Vicuna-7B), where most facts in MQuAKE dataset are in the two pre-trained models. As for the Eval-DU, we test with models fine-tuned on the synthetic dataset, which allows us to have more flexible choices of the testing LLMs. We choose four popular LLMs to evaluate on Eval-DU (Phi-1.5, GPT2-XL, Llama2-7b, Llama3-8b).
Across both benchmarks, we find that none of the methods achieves a Success-DU score above 0.8 while maintaining an accuracy of at least 80\%, highlighting the difficulty of effective deep unlearning.
Furthermore, the gap between deep unlearning and superficial unlearning is not neglectable, it is a minimum of x across all models; it is especially pronounced on Eval-DU, where the deduction chains towards target facts are longer and more complex. 

This illustrates that the machine unlearning methods of today are largely  inadequate for robustly unlearning facts from LLMs.
We hypothesize that this might be because the existing unlearning methods do not sufficiently account for the nature of facts and the deductions among them. 
We posit that future methods that unlearn facts from LLMs should be aware of the deductions between facts.


%% file: sections/preliminary.tex
\section{Representing and Reasoning about Factual Knowledge in LLMs}

\paragraph{Representing factual knowledge with triplets.} Despite the impressive factual recall capabilities of large language models (LLMs), the knowledge they encode is typically unstructured and not explicitly represented. In this work, we adopt the formalism of knowledge bases~\citep{nickel2015review, bordes2013translating, toutanova2015observed, hogan2021knowledge} (i.e. knowledge graphs) to model factual knowledge stored in LLMs. This representation has been extensively studied in the literature, including in knowledge graph embeddings~\citep{bordes2013translating,yang2015embedding,8047276}, neuro-symbolic reasoning~\citep{rocktaschel2017end, zhang2021neural, xu2022ruleformer,cheng2023neural,luo2023chatrule}, and their downstream applications knowledge completion. More recently, it has also been used to study knowledge editing~\citep{meng2022locating, meng2023massediting} and unlearning~\citep{wang2024large,choi2024breaking} in LLMs, as well as in systems that integrate such structured knowledge with LLMs~\citep{sun2024thinkongraph,jiang2023structgpt,baek2023knowledge}.

Formally, given a set of objects $\calO$ and relations $\calT$, a fact $k$ is represented as a triplet $(o_1, r, o_2)$, where $r \in \calT$ and $o_1, o_2 \in \calO$. For example, the fact “Camila Flores's child is Wyatt Ross” is represented as $(\textit{Camila Flores}, \textit{child}, \textit{Wyatt Ross})$. A knowledge base $\calK$ is then defined as a set of such triplets: $\calK \subseteq \calO \times \calT \times \calO$.

\paragraph{Modeling deductive reasoning with logical rules.} To capture the adversary’s ability to infer new facts from known ones, we model the deduction process using logical rules~\citep{lloyd2012foundations,muggleton1994inductive}. 
They are widely studied for new knowledge discovery~\citep{galarraga2013amie,yang2017differentiable,xu2022ruleformer,cheng2023neural,luo2023chatrule} and play a central role in other directions such as formal verification~\citep{lloyd2012foundations,bjorner2015horn}.

A logical rule $R$ has the form $B_1 \wedge \cdots \wedge B_n \rightarrow A$, where $B_1, \dots, B_n$ and $A$ are atoms, each represented as a triplet $(X, r, Y)$ with logical variables $X, Y$ and relation $r$. For example, (X, \textit{husband}, Z)$\wedge$ (Y, \textit{father}, Z)  $\to$ (X, \textit{child}, Y) states that if $X$ is the husband of $Z$, and $Y$ is the father of $Z$, then $X$ is the child of $Y$. By grounding the logical variables with objects from $\calO$, the body of the rule ($B_1 \wedge \cdots \wedge B_n$) can be used to deduce the head fact $A$.

\paragraph{Alternative knowledge and deduction formalisms.}
Knowledge representations can range from formal symbolic structures such as n-ary relations~\citep{w3c2006nary} (generalizing triplets) to more expressive, informal formats like natural language. Correspondingly, deduction mechanisms can vary from symbolic logic to neural inference over text (e.g. through LLMs~\citep{wei2022chain}), in the deterministic or probablistic way~\citep{manhaeve2018deepproblog,qu2019probabilistic}. Extending our framework of deep unlearning to these broader knowledge representations and reasoning paradigms is an important direction for future work.

%% file: sections/formulation.tex
\section{Deep Unlearning}
Prior work in fact unlearning from LLMs focuses on simply unlearning the target fact in isolation. 
This might cause the LLM to forget only this one specific fact, but retain others that can be combined to deduce back the target fact. 
In this section, we introduce the new setting of unlearning, \textit{\problemname}, which considers such logical deductions.

\subsection{Fact \problemname\ }
\begin{figure*}
\centering
	\includegraphics[width=\linewidth]{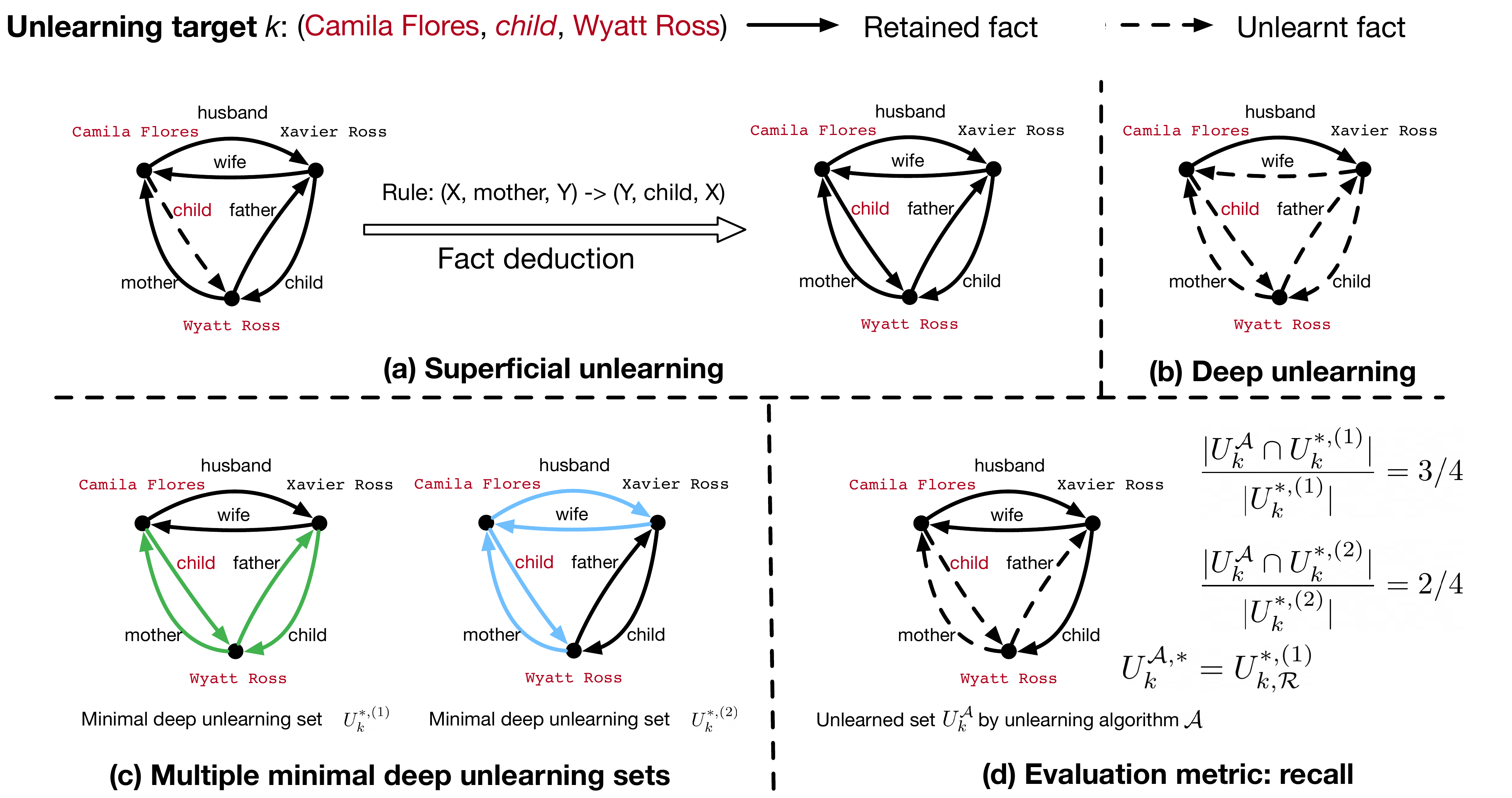}
	\caption{An illustration of \textit{\problemname}. (a) an example of superficial unlearning; (b) an example of \problemname; (c) two different minimal \problemname\ sets for unlearning the same target fact; (d) the calculation of our proposed evaluation metric \textit{recall}.}
	\label{fig:illustration}
\end{figure*}

Given the knowledge base $\calK$ of an LLM prior to unlearning, let $U_k^{\calA} \subseteq \calK$ denote the set of facts removed by an unlearning method $\calA$ when attempting to unlearn a target fact $k$. To achieve \problemname, we require that the target fact $k$ is no longer inferable from the retained knowledge $\calK \setminus U_k^{\calA}$ under the given rule set $\calR$.
In other words, even if an adversary has access to the LLM's remaining facts after unlearning and can apply any rule in $\calR$, they should be unable to reconstruct or deduce the target fact.

This notion relies on the concept of \textit{deductive closure}~\citep{cheney2009provenance,cohen2016tensorlog,huang2021scallop}, a standard construct in logical reasoning. The deductive closure of a knowledge base $\calK$ with respect to a rule set $\calR$ includes all facts that can be derived by applying any rule in $\calR$ to the facts in $\calK$.
For example, suppose the knowledge base $\calK$ contains the facts
$(\textit{Camila Flores}, \textit{husband}, \textit{Xavier Ross})$ and
$(\textit{Xavier Ross}, \textit{child}, \textit{Wyatt Ross})$,
and the rule set $\calR$ includes (X, \textit{husband}, Z)$\wedge$ (Y, \textit{father}, Z)  $\to$ (X, \textit{child}, Y).
Then the deductive closure of $\calK$ must contain the fact $(\textit{Camila Flores}, \textit{child}, \textit{Wyatt Ross})$.
We formally state the deductive closure as follows:
\begin{definition}[Deductive closure]
	The deductive closure of knowledge base $\calK$ with respect to the rule set $\calR$, denoted as $\Omega(\calK, \calR)$, is the smallest set such that (1) $\calK\subseteq \Omega(\calK, \calR)$; (2) $\Omega(\calK, \calR)$ is \textit{deductively closed} with respect to $\calR$.
\end{definition}

We now define the central concept, \textbf{deep unlearning}, of our work:
\begin{definition}[\problemnamecap]
	The unlearning method $\calA$ \problemnamethird\ the fact $k$ with respect to the rule set $\calR$ if the fact $k$ is not in the deductive closure $\Omega(\calK\backslash U_k^{\calA}, \calR)$.
\end{definition}
We refer to unlearning outcomes where the target fact is successfully removed from the model but remains deducible from retained knowledge as \textit{superficial} unlearning; an example is illustrated in Figure~\ref{fig:illustration}(a).
In contrast, Figure~\ref{fig:illustration}(b) depicts a successful case of \problemname, where only the fact $(\textit{Camila Flores}, \textit{husband}, \textit{Xavier Ross})$ is retained, and the target fact is no longer derivable under any rule in $\calR$.

Notably, in this example of \problemname\ (Figure~\ref{fig:illustration}(b)), the unlearning would still be valid even if the fact $(\textit{Xavier Ross}, \textit{wife}, \textit{Camila Flores})$ were retained. This highlights an important consideration: in practice, we prefer unlearning strategies that remove as few facts as necessary.
This motivates our next definition of \textbf{minimal \problemname}.

\begin{definition}[Minimal \problemname]
\label{def:m_du}
	Given a fact $k$, the minimal \problemname\  set $U_{k}^{*}$ to unlearn the fact $k$ w.r.t. the rule set $\calR$ should meet two requirements: (1) $k\notin \Omega(\calK\backslash U_{k}^{*}, \calR)$, (2) $\forall U\subset U_{k}^*$, $k\in \Omega(\calK\backslash U, \calR)$.
	Moreover, the unlearning method $\calA$ minimally \problemnamethird\ $k$  w.r.t. $\calR$ if $U_k^{\calA}$, the set of facts that is removed by $\calA$ for unlearning $k$, is a minimal \problemname\  set.
\end{definition}
Minimal \problemname\ refers to the smallest set of facts that must be removed from the knowledge base to ensure the target fact is no longer deducible. Importantly, such minimal sets are not necessarily unique: Figure~\ref{fig:illustration}(c) gives an example. 
This non-uniqueness highlights the flexibility in unlearning strategies, which the evaluation metric should account for.



\subsection{Connection to the existing unlearning metrics and problems}
\label{sec:connection}
\paragraph{Connection to other metrics of fact unlearning.}
Recent work has introduced various metrics to evaluate the effectiveness of fact unlearning in language models~\citep{tofu2024,li2024the,yao2024machine,patil2024can,lucki2024adversarial,joshi2024towards}. These metrics primarily assess whether the target LLM continues to retain or recall the target fact. A common approach is to directly query the model with prompts related to the fact and check whether its responses imply retention. For example, \citet{patil2024can} extend this by using paraphrased queries to more robustly probe for the fact's presence. We refer readers to Section~\ref{sec:related_work} for additional details about other work. \textbf{Our work takes a complementary perspective}: rather than only measuring how much of a target fact is retained, we ask what additional related facts must be removed to ensure the fact is no longer inferable. This question forms the core motivation behind our notion of deep unlearning.

\paragraph{Connection to multi-hop fact unlearning.}
A concurrent work by \citet{choi2024breaking} explores multi-hop unlearning, which also involves reasoning-based fact deduction. However, their formulation fundamentally differs from our concept of deep unlearning. Consider a multi-hop reasoning instance:
$$
(o_1, r_1, o_2) \wedge (o_2, r_2, o_3) \to (o_1, \overline{r_1r_2}, o_3),
$$
where all three facts are present in the LLM’s knowledge base. In their setting, unlearning the single-hop fact $(o_1, r_1, o_2)$ should also eliminate its logical implication $(o_1, \overline{r_1r_2}, o_3)$ from the model. In contrast, our notion of deep unlearning is defined in the opposite direction: to unlearn the inferred fact $(o_1, \overline{r_1r_2}, o_3)$, one must also unlearn at least one supporting fact -- either $(o_1, r_1, o_2)$ or $(o_2, r_2, o_3)$.  \textbf{This reflects a fundamental difference how the reasoning is treated between deep unlearning and multi-hop unlearning}: deep unlearning focuses on preventing adversarial reconstruction of the target fact by removing the supporting facts that can be used to deduce it, where the adversary's reasoning capabilities are considered for such deductions; in contrast, multi-hop unlearning aligns with multi-hop knowledge editing~\citep{zhong2023mquake}, targeting the downstream implications of a fact, where reasoning is used to propagate the unlearning effects.

\section{Evaluation Metrics for Deep Unlearning}
We propose three metrics for deep unlearning to evaluate an unlearning method $\calA$: \textit{Success-DU} and \textit{Recall} for measurining the unlearning utility and \textit{Accuracy} for measuring retaining utility. 

\textbf{Success-DU} directly follows the definition of deep unlearning. Let $U_k^{\calA}$ denote the set of facts removed from the LLM by method $\calA$ when unlearning target fact $k$. Success-DU evaluates whether $\calA$ successfully achieves deep unlearning by checking if $k$ can no longer be inferred from the remaining knowledge base:
\begin{equation}
\label{eq:sucess_du}
\text{Success-DU}(\calA, k; \calK, \calR)=	\mathds{1}\left[ k\notin \Omega(\calK\backslash U_{k}^{\calA}, \calR)\right],
\end{equation}
where $\Omega(\cdot, \calR)$ denotes the deductive closure under rule set $\calR$.

While Success-DU captures a binary outcome, \textbf{Recall} is to measure the degree of \problemname\   achieved by the unlearning method $\calA$, offering a more detailed view of $\calA$’s performance.
Specifically, it calculates the percentage of any minimal \problemname\  set that has been unlearnt by the method $\calA$.
Because the minimal \problemname\  set is not unique, the recall is defined with the minimal \problemname\  set that $U_{k}^{\calA}$ (the set of facts removed by $\calA$ for unlearning the fact $k$) overlaps the most.
Formally, let $\calM_{k, \calR, \calK}$ denote the set of all minimal \problemname\  sets to unlearn $k$ (from the knowledge base $\calK$ w.r.t. the rule set $\calR$). 
The recall for a given unlearning method $\calA$ to unlearn $k$ is defined as
\begin{equation}
\label{eq:recall}
\text{Recall}(\calA, k; \calK, \calR)=	\max_{U_{k}^*\in \calM_{k, \calR, \calK}}\frac{|U_{k}^{\calA}\cap U_{k}^*|}{|U_{k}^*|}.
\end{equation}
We also denote with $U_{k}^{\calA, *}$ the minimal \problemname\  set that $U_{k}^{\calA}$ overlaps the most, which is used for calculating the recall, $U_{k}^{\calA, *}:=\arg\max_{U_{k}^*\in \calM_{k, \calR, \calK}}\frac{|U_{k}^{\calA}\cap U_{k}^*|}{|U_{k}^*|}$.
Figure~\ref{fig:illustration}(d) shows an example of calculating this recall. 
There are two minimal \problemname\  sets for unlearning the target fact. By definition $U_{k}^{\calA, *}=U_{k}^{*, (1)}$ is picked for the recall value.

Note that unlike Success-DU, recall is a continuous measure that reflects how effectively the unlearning method $\mathcal{A}$ achieves deep unlearning. From the adversary’s perspective, a lower recall value indicates less difficulty in recovering the target fact. This is because more candidate fact combinations remain in the model, enabling the adversary to probe these alternatives and more easily infer the specific target fact.

\textbf{Accuracy} is defined to measure utility of the LLM.
We calculate the accuracy on the knowledge base after excluding the minimal \problemname\  set (for calculating the recall), $\calK\backslash U_{k}^{\calA,*}$, :
\begin{equation}
\label{eq:acc}
\text{Accuracy}(\calA, k; \calK, \calR)=\frac{|(\calK\backslash U_{k}^{\calA, *})\backslash U_{k}^{\calA})|}{|\calK\backslash U_{k}^{\calA, *}|}.
\end{equation}
Ideally when the unlearning method $\calA$ exactly unlearns a \problemname\ set, both recall and accuracy are $1$; otherwise, either the unlearning method does not \problemnameactive\ the target fact $k$ (recall$<1$), or it unlearns extraneous facts for unlearning $k$ (accuracy$<1$). 


\begin{algorithm}[!t]
\caption{MDUS$(k, \calK, \calR; N_{\rm seed})$ -- Generating multiple \textit{M}inimal \textit{D}eep \textit{U}nlearning \textit{S}ets}
\textbf{Input:} The target fact $k$, the knowledge base $\calK$, the rule set $\calR$, the number of seeds $N_{\rm seed}.$
	\begin{algorithmic}[1]
		\STATE $\hat{\calM}_{k, \calR, \calK}=\{\}$.
		\FOR{$n_{\rm seed}=1, \cdots, N_{\rm seed}$}
		\STATE $U_k=$DUS$(k, \calK, \calR)$.\ \ $\backslash\backslash$ Algorithm~\ref{algo:gen_sus}
		\STATE $U_k^*$=RP$(k, \calK, \calR, U_k)$.  \ \ $\backslash\backslash$ Algorithm~\ref{algo:pr_msus}
		\STATE $\hat{\calM}_{k, \calR, \calK} = \hat{\calM}_{k, \calR, \calK}\cup \{U_k^*\}$.
		\ENDFOR
\STATE \textbf{Output: $\hat{\calM}_{k, \calR, \calK}$}
	\end{algorithmic}
	\label{algo:main}
\end{algorithm}

\begin{algorithm}[!t]
\caption{DUS$(k, \calK, \calR)$ -- Random generation of the \textit{D}eep \textit{U}nlearning \textit{S}et}
\textbf{Input:} The target fact $k$, the knowledge base $\calK$, the rule set $\calR.$
	\begin{algorithmic}[1]
		\STATE $\hat{U}_k=\{k\}$, $T=\{k\}$
		\WHILE{$T\neq \emptyset$}
		\STATE Uniformly randomly pick $k_{\rm cur}\in T$. $T = T\backslash \{k_{\rm cur}\}$
		\STATE Find all initializations of rules $\calI_{k_{\rm cur}}$ that implies $k_{\rm cur}$ and denote the size $|\calI_{k_{\rm cur}}|$ as $m_{k_{\rm cur}}$:
		$$\calI_{k_{\rm cur}} = \{I_j|\forall j\in[m_{k_{\rm cur}}],  $$
		$$I_j=(k_1^j, \cdots, k_{n_j}^j, k_{\rm cur})\in \Omega(\calK, \calR)\times \cdots \times \Omega(\calK, \calR)$$
		$$\text{ is an initiation of the rule } B_1^j\wedge \cdots\wedge B_{n_j}^j\to A_j\in \calR \}$$
		\FOR{$(k_1^j, \cdots, k_{n_j}^j, k_{\rm cur})\in \calI_{k_{\rm cur}}$ and $\{k_1^j, \cdots, k_{n_j}^j\}\cap \hat{U}_k=\emptyset$ in a random order}
			\STATE Uniformly randomly pick $k^j$ from $\{k_1^j, \cdots, k_{n_j}^j\}$. $\hat{U}_k=\hat{U}_k\cup \{k^j\}$, $T=T\cup \{k^j\}$.
		\ENDFOR
		\ENDWHILE
		\STATE \textbf{Output: }$U_k:=\hat{U}\cap \calK$.
	\end{algorithmic}
	\label{algo:gen_sus}
\end{algorithm}


\subsection{Approximation Algorithm for Calculating Recall and Accuracy} 
\label{sec:app_algo_1}
Calculating both recall and accuracy rely on solving an optimization problem
$$
U_{k}^{\calA, *}:=\arg\max_{U_{k}^*\in \calM_{k, \calR, \calK}}\frac{|U_{k}^{\calA}\cap U_{k}^*|}{|U_{k}^*|},
$$
where $\calM_{k, \calR, \calK}$ denote the set of all minimal \problemname\  sets to unlearn $k$ (from the knowledge base $\calK$ with respective to the rule set $\calR$).
However, finding the exact $U_{k}^{\calA, *}$ in general can be NP-hard~\citep{Skiena2020-mw}. 
Alternatively, we propose Algorithm~\ref{algo:main}, which is able to find multiple minimal \problemname\  sets $\hat{\calM}_{k, \calR, \calK}$.
Then it is efficient to find $\hat{U}_{k}^{\calA, *}:=\arg\max_{U_{k}^*\in \hat{\calM}_{k, \calR, \calK}}\frac{|U_{k}^{\calA}\cap U_{k}^*|}{|U_{k}^*|}$ and approximately calculate the recall and accuracy afterwards.

The idea in Algorithm~\ref{algo:main} is to generate a single minimal \problemname\ set with some randomness (line 3-4) and to repeat this generation process to attain multiple minimal \problemname\ sets.
There are two steps to find a single minimal \problemname\  set;
\begin{algorithm}[!t]
\caption{RP$(k, \calK, \calR, U_k)$ -- \textit{R}andom \textit{P}runing the \problemname\  set}
\textbf{Input:} The target fact $k$, the knowledge base $\calK$, the rule set $\calR$, the \problemname\  set $U_k$%
	\begin{algorithmic}[1]
		\STATE $C=\{\}$, $t=0$, $U_k^*=U_k.$
		\WHILE{$C\neq\emptyset$ or $t=0$}
            \STATE $C=\{\}$, $t=t+1$
		\FOR{$k_{\rm cur}$ in randomly shuffled $U_k^*$}
		\IF{$k\notin \Omega(\calK\backslash (U_k^* \backslash  \{k_{\rm cur}\}), \calR)$}
			\STATE $C=C\cup \{k_{\rm cur}\}$, $U_k^* = U_k^*\backslash\{k_{\rm cur}\}$
		\ENDIF
		\ENDFOR
		\ENDWHILE
		\STATE \textbf{Output: }$U_k^*$.
	\end{algorithmic}
	\label{algo:pr_msus}
\end{algorithm}
\begin{enumerate}[leftmargin=*,nosep]
\item Find any \problemname\  set  (Algorithm~\ref{algo:gen_sus}). We enumerate the rules and find all combinations of facts that can imply fact $k$ (line 4).
For each combination, if no facts in this combination are in the returning set $U_k$, we randomly pick one fact from this combination and add it to the returning set $U_k$ (lines 5-7). 
Additionally, for the picked fact in any combination, we repeat the above process but for this fact recursively.
This algorithm guarantees that fact $k\notin \Omega(\calK\backslash U_k, \calR)$ and randomness from picking fact in each combination and the order for going through the combinations brings diversity in the results.
\item Prune $U_k$, a \problemname\  set, to a minimal \problemname\  set $U_k^*$ (Algorithm~\ref{algo:pr_msus}). We go through every fact $k_{\rm cur}$ in $U_k$ one by one and check if $U_k\backslash\{k_{\rm cur}\}$ from $\calK$ is still a \problemname\ set. If yes, we can safely remove $k_{\rm cur}$ from current $U_k$ and repeat this process until there is no $k_{\rm cur}\in U_k$ that can be removed. The $U_k^*$ returned by this algorithm is guaranteed to be a minimal \problemname\ set, and the randomness in the order of checking $k_{\rm cur}\in U_k$ brings diversity in the results.
\end{enumerate}

\paragraph{Theoretical guarantee of Algorithm~\ref{algo:main}} We are going to state following guarantee of Algorithm~\ref{algo:main}.
\begin{theorem}
$\hat{M}_{k, \calR, \calK}$ returned by Algorithm~\ref{algo:main} is a collection of minimal \problemname\ sets.	
\end{theorem}

\begin{proof}
We can first prove $k\notin \Omega (\calK\backslash U_k, \calR)$, where $U_k$ at line 3 in Algorithm~\ref{algo:main} is returned by Algorithm~\ref{algo:gen_sus}. The proof has two steps:
\begin{enumerate}
\item 	We can have $\Omega( \Omega(\calK, \calR)\backslash \hat{U}_k, \calR  ) = \Omega(\calK, \calR)\backslash \hat{U}_k$, where $\hat{U}_k$ here is the $\hat{U}_k$ after line 8 in Algorithm~\ref{algo:gen_sus}. Otherwise, by the definition of deductive closure, there exists $k'\notin \Omega(\calK, \calR)\backslash \hat{U}_k$ and $k'$ can be deduced from initiation of the rule where all facts on the left of the rule are in $\Omega(\calK, \calR)\backslash \hat{U}_k$, i.e. not in $\hat{U}_k$. However, this can be a contradiction because if $k'\notin \Omega(\calK, \calR)\backslash \hat{U}_k$, $k'$ must be in $\hat{U}_k$ and line 5-7 in Algorithm~\ref{algo:gen_sus} can guarantee that for any initiation of any rule that can imply $k'$, there is at least one fact on the left of the rule in $\hat{U}_k$.
\item From line 1 in Algorithm~\ref{algo:gen_sus}, we know that $k\in \hat{U}_k$. This means that $k\notin \Omega(\calK, \calR)\backslash \hat{U}_k = \Omega( \Omega(\calK, \calR)\backslash \hat{U}_k, \calR  ) $, where the equality is from step 1. On the other hand, $(\calK\backslash U_k) = (\calK\backslash \hat{U}_k) \subseteq \Omega(\calK, \calR)\backslash \hat{U}_k)$ where the equality comes from the definition $U_k = \calK\cap \hat{U}_k$ at line 9 in Algorithm~\ref{algo:gen_sus}. $k\notin \Omega( \Omega(\calK, \calR)\backslash \hat{U}_k, \calR  )$ and $(\calK\backslash U_k)\subseteq \Omega(\calK, \calR)\backslash \hat{U}_k)$ together imply $k\notin \Omega (\calK\backslash U_k, \calR)$.
\end{enumerate}

We now have $k\notin \Omega (\calK\backslash U_k, \calR)$, then we are going to prove $U_k^*$ returned by Algorithm~\ref{algo:pr_msus} is a minimal \problemname\  set. 
From Algorithm~\ref{algo:pr_msus}, it is obvious that $k\notin  \Omega(\calK\backslash U_k^*, \calR)$. If it is not the minimal \problemname\  set, then there exists $U'\subset U_k^*$ s.t. $k\notin  \Omega(\calK\backslash U, \calR)$ and there is an $k'$ s.t. $k'\notin U_k^*$ and $k'\in U'$. However, this is a contradiction, because Algorithm~\ref{algo:pr_msus} only returns $U_k^*$ if $\forall k' \notin U_k^*$, $k\in \Omega(\calK\backslash  U_k^*\backslash \{k'\}, \calR)$.

Now we can conclude $U_k^*$ at line 4 in Algorithm~\ref{algo:main} is a minimal \problemname\  set, and our proof is done.
\end{proof}

\begin{figure}[!t]
      \centering
      \includegraphics[width=\linewidth]{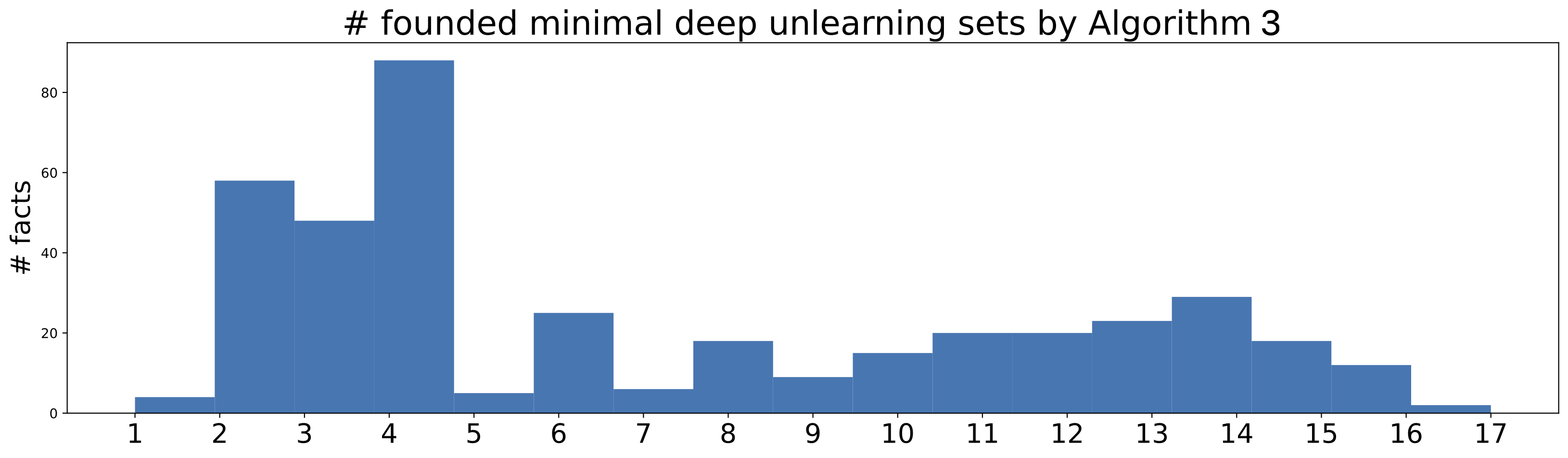}
\caption{Histogram of \# minimal \problemname\ sets founded by Algorithm~\ref{algo:main}.}
\label{fig:num_min_dist}
\end{figure}

\begin{figure}[!t]
      \centering
      \includegraphics[width=\linewidth]{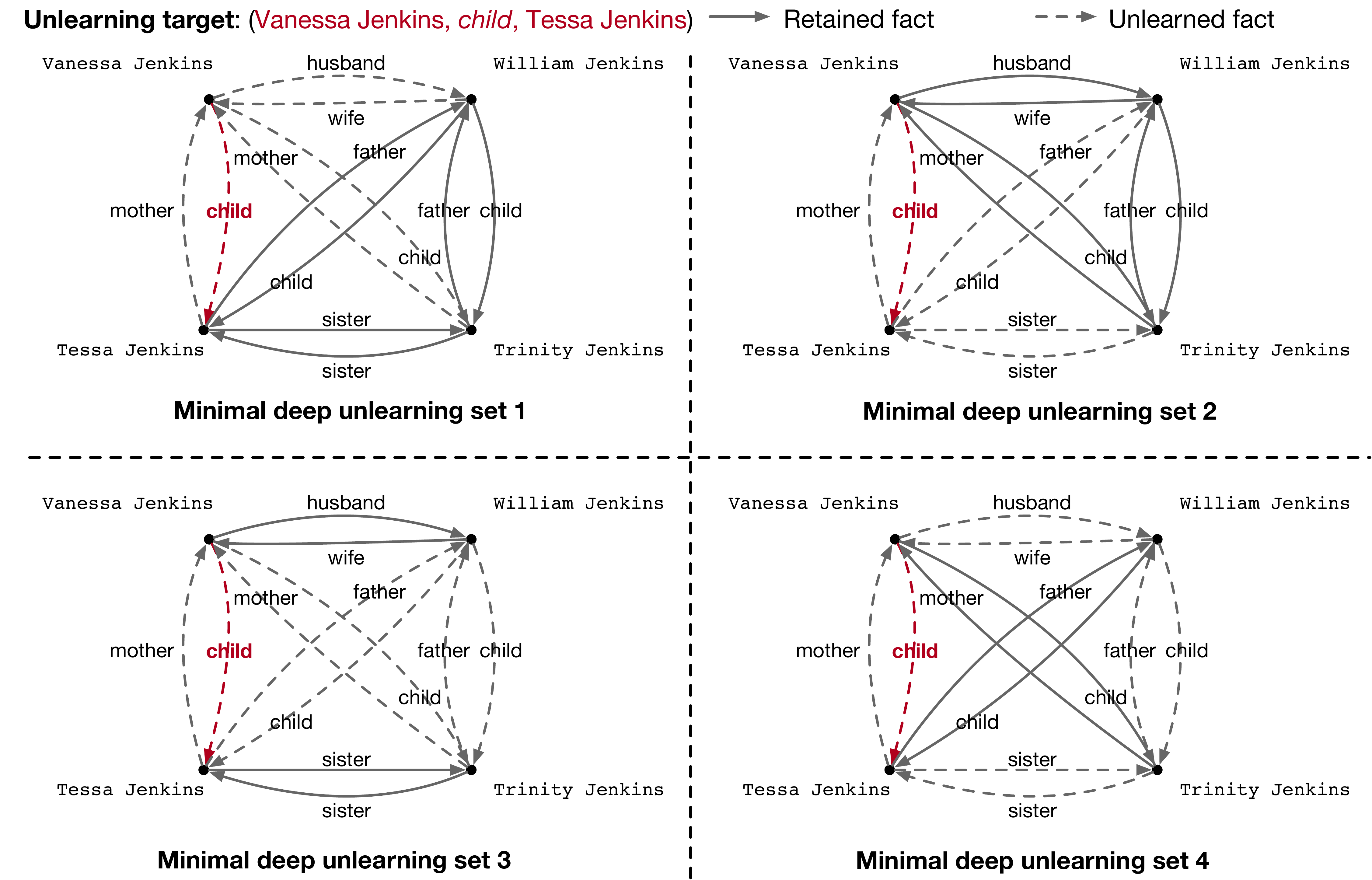}
\caption{An example of 4 minimal \problemname\ sets founded by Algorithm~\ref{algo:main}.}
\label{fig:example_algo1}
\end{figure}

\paragraph{Empirical Evaluation with Algorithm~\ref{algo:main}}
By running Algorithm~\ref{algo:main} on the facts in the synthetic dataset introduced in the later section, we find that Algorithm~\ref{algo:main} is capable of generating a diverse set of minimal \problemname\ sets.
For more than half of the facts in our synthetic dataset, Algorithm~\ref{algo:main} can return 6-17 different minimal \problemname\ sets; Figure~\ref{fig:num_min_dist} shows the histogram of \# minimal \problemname\ sets founded by Algorithm~\ref{algo:main}.
This demonstrates the effectiveness of Algorithm~\ref{algo:main} and hence leads to a good approximation for computing the recall in Equation~\ref{eq:recall}.
Figure~\ref{fig:example_algo1} shows an example of 4 minimal \problemname\ sets founded by Algorithm~\ref{algo:main}.

%% file: sections/synthetic_dataset.tex
\section{Datasets for Evaluating Deep Unlearning}
\label{sec:synth_data}
To systematically evaluate deep unlearning in large language models (LLMs), we construct and leverage two complementary datasets, both specifying unlearning targets together with the associated facts and logical rules that can be used to deduce them. 
The first is based on real-world factual knowledge extracted from pretrained LLMs from the literature, where deep unlearning involves blocking a single-step deduction.
The second is our newly constructed \emph{semi-synthetic} dataset, designed to provide complete observability and more general logical rules, where multi-step deductions are involved in deep unlearning.
These datasets allow us to assess the effectiveness of deep unlearning methods under both realistic and controlled knowledge settings. 

%

\subsection{Evaluating deep unlearning with real-world knowledge (MQuAKE)}
To evaluate deep unlearning in the context of real-world factual knowledge, we leverage the MQuAKE dataset~\citep{{zhong2023mquake}}, which provides multiple multi-hop reasoning instances in the form:
$$
(o_1, r_1, o_2) \wedge \cdots \wedge (o_{h}, r_h, o_{h+1}) \to (o_1, \overline{r_1\cdots r_h}, o_{h+1}),
$$
where $\overline{r_1\cdots r_h}$ denotes a super-relation inferred from the composed single-hop relations.
We focus on unlearning the super-relation fact $(o_1, \overline{r_1\cdots r_h}, o_{h+1})$.
The corresponding minimal deep unlearning set by Definition~\ref{def:m_du} has always size of two: the target super-relation fact and one arbitrary supporting single-hop fact from its multi-hop chain.

\subsection{\textit{Eval}uating \textit{D}eep \textit{U}nlearning with fully controlled synthetic knowledge (Eval-DU)}
While MQuAKE enables the evaluation of deep unlearning through multi-hop reasoning for the real-world facts, it has two limitations.
First, when considering the minimal deep unlearning set, only a single deduction step through the multi-hop rule is involved. 
In practice, however, deep unlearning may involve more diverse logical rules and require reasoning over multiple deductive steps.
Moreover, MQuAKE only provides a partial view of the model’s internal knowledge. This limited observability might lead to unreliable conclusions about unlearning effectiveness.\footnote{For instance, a model may retain some supporting facts that still imply the unlearned target, yet appear successful in evaluation due to their absence from the probed knowledge base.} In fact, this limitation is inherent to any evaluation using real-world knowledge extracted from pretrained models.

To address these challenges, we construct a \emph{semi-synthetic} dataset, \dataname, which enables controlled evaluations of deep unlearning under more complex logical rules and complete knowledge observability. We evaluate deep unlearning on models fine-tuned with this dataset.
We locate our dataset in a family network, which is a common scenario to study rule mining and knowledge discovery in the literature~\citep{galarraga2013amie,cheng2023neural,luo2023chatrule}.
This semi-synthetic dataset includes a \emph{synthetic} knowledge base consisting of $400$ family relationships and $300$ biographical facts among $100$ fictitious people, as well as a set of \emph{realistic} logical rules, which are deductions among family relationships.

Family relationships include \textit{child}, \textit{father}, \textit{mother}, \textit{husband}, \textit{wife}, \textit{brother}, \textit{sister}, \textit{aunt}, \textit{uncle}, \textit{nephew}, \textit{niece}.
Biographies include \textit{birthyear}, \textit{birthplace}, and \textit{job}.
Table~\ref{tab:knowledge_examples} shows some examples of facts in family relationships and biographies, together with the question-answer pairs for checking whether this fact is in the LLM or not.
Moreover, the rule set $\calR$ has 48 rules, which are used to deduce the facts in family relationships.
Table~\ref{tab:rule_examples} shows all rules that can imply the fact that has \textit{child} as the relationship.

The construction of Eval-DU provides two key advantages for evaluating deep unlearning.
First, deep unlearning on Eval-DU presents a significantly greater challenge, requiring the removal of more interdependent facts than MQuAKE.
In Eval-DU, the size of the minimal deep unlearning set for a given target fact ranges from 1 to 17, reflecting the presence of multi-step deduction chains; Figure~\ref{fig:num_min_dist} in the appendix shows the detailed distribution of sizes in Eval-DU. This contrasts with MQuAKE, where the deduction depth is limited and the minimal set size is fixed at 2. 
Second, because Eval-DU operates in a fully synthetic space, the entire knowledge base is observable. This complete visibility enables more accurate and reliable evaluation of deep unlearning outcomes.


\begin{table*}[!t]
\centering
\caption{Examples of synthetic facts in family relationships and biographies.}
\begin{tabular}{c|cc}
	\toprule
	Fact & Question & Answer\\
	\midrule
	 	(Reid Perry, \textit{father}, Richard Perry) & Who is Richard Perry to Reid Perry? & Father\\
		(Richard Perry, \textit{child}, Quentin Perry) & Who is Quentin Perry to Richard Perry? & Child\\
		(Quinn Gray, \textit{sister}, Rachel Gray) & Who is Rachel Gray to Quinn Gray? & Sister\\
	\midrule
	 	(Sloane Lee, \textit{birthyear}, 1908) & What is the birth year of Sloane Lee? & 1908\\
		(Sloane Lee, \textit{birthplace}, Washington state) & What is the birthplace of Sloane Lee? & Washington state\\
		(Sloane Lee, \textit{job}, Banker) &  What is the job of Sloane Lee? & Banker\\
	\bottomrule
\end{tabular}	
\label{tab:knowledge_examples}
\end{table*}

\begin{table*}[!t]
\centering
\caption{Rules that deduce any fact having \textit{child} as relation.}
\begin{tabular}{cc}
	\toprule
	(B, \textit{mother}, A) $\to$ (A, \textit{child}, B) &
(B, \textit{father}, A) $\to$ (A, \textit{child}, B)\\
(C, \textit{mother}, A) $\wedge$ (B, \textit{brother}, C) $\to$ (A, \textit{child}, B) & (C, \textit{mother}, A) $\wedge$ (B, \textit{sister}, C) $\to$ (A, \textit{child}, B)
\\
(C, \textit{father}, A) $\wedge$ (B, \textit{sister}, C) $\to$ (A, \textit{child}, B) &
(C, \textit{father}, A) $\wedge$ (B, \textit{brother}, C) $\to$ (A, \textit{child}, B)\\
(A, \textit{child}, C) $\wedge$ (B, \textit{sister}, C) $\to$ (A, \textit{child}, B) &
(A, \textit{child}, C) $\wedge$ (B, \textit{brother}, C) $\to$ (A, \textit{child}, B)\\
(A, \textit{child}, C) $\wedge$ (B, \textit{wife}, C) $\to$ (A, \textit{child}, B) &
(A, \textit{child}, C) $\wedge$ (B, \textit{husband}, C) $\to$ (A, \textit{child}, B)\\
	\bottomrule
\end{tabular}
\label{tab:rule_examples}
\end{table*}

To better approximate a realistic knowledge base, we incorporate several design choices in generating the synthetic family network:
\begin{itemize}
	\item \textit{Family network generation.} We recursively expand the network. Given a node (person), with a certain probability, we generate the parents, spouse, and children of this person. We control the whole family network in 4 generations. The number of children from any couple is sampled from a truncated ($\leq 4$) geometric distribution. 
	\item \textit{Name generation.} We collect two lists of first names for males and females separately and assign the first name to each person according to gender. As for the last name, each person's last name is the same as the father's if the father exists in the network. There is only one special case where the female's last name has a small probability of switching to her husband's.
	\item \textit{Biography generation.} We have three biographical attributes, birth year, birthplace, and job:
		\begin{itemize}[leftmargin=*,nosep]
		\item The birth years of people are aligned with their relationships. The birth year of any child is from a truncated Gaussian distribution given his/her mother's birth year. The difference in birth years of a couple is sampled from a reasonable distribution as well.
		\item The birthplace is the state in the United States. The child's birthplace is the same as the birthplace of the parent with a high chance, or sampled from the population distribution in the United States.
		\item The job list is collected from GPT4 for every ten years in 1900-2020. The job of each person is picked based on the birth year.
		\end{itemize}		 
\end{itemize}
These realistic design choices narrow the gap between synthetic and real-world unlearning tasks, enabling controlled yet meaningful evaluation. 

\begin{figure}[!t]
      \centering
      \includegraphics[width=\linewidth]{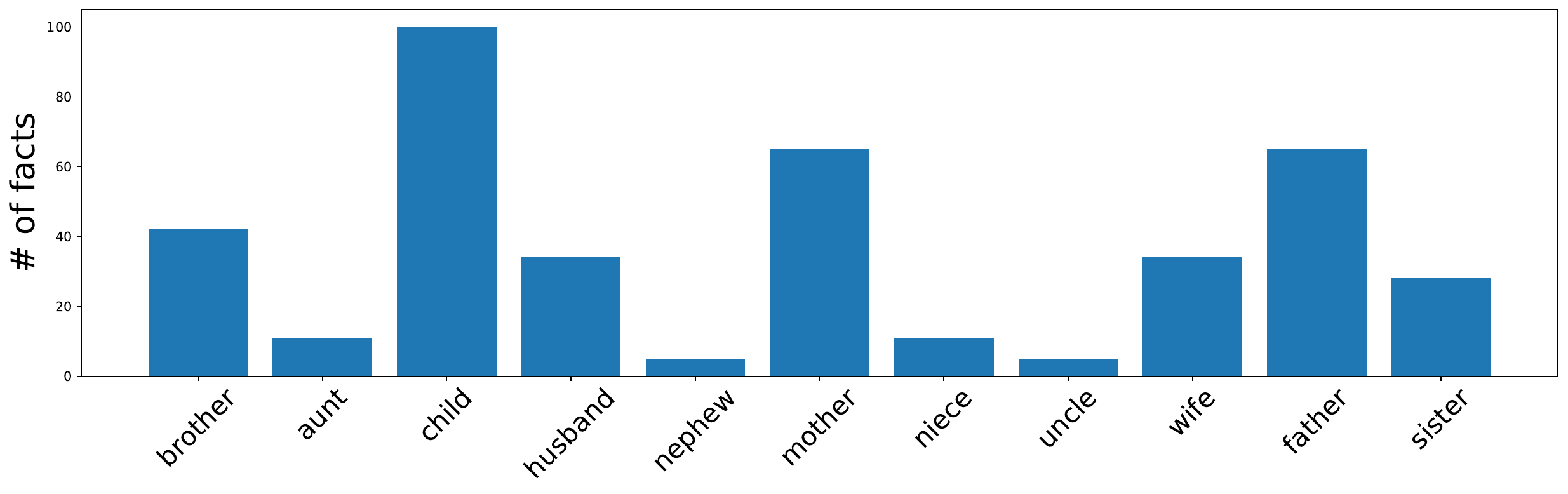}
\vspace{-4ex}
\caption{Distribution of relations in our synthetic dataset.}
\label{fig:relation_dist}
\end{figure}

\begin{figure}[!t]
      \centering
      \includegraphics[width=\linewidth]{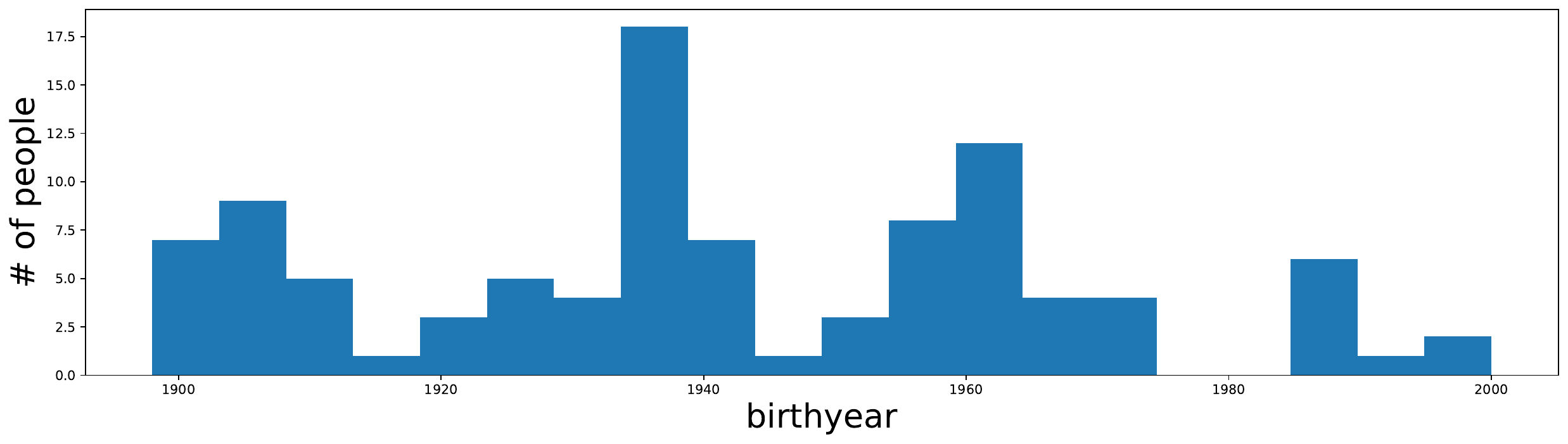}
      \vspace{-4ex}
\caption{Distribution of birth years of fictitious people in our synthetic dataset.}
\label{fig:birthyear_dist}
\end{figure}

\begin{figure}[!t]
      \centering
      \includegraphics[width=\linewidth]{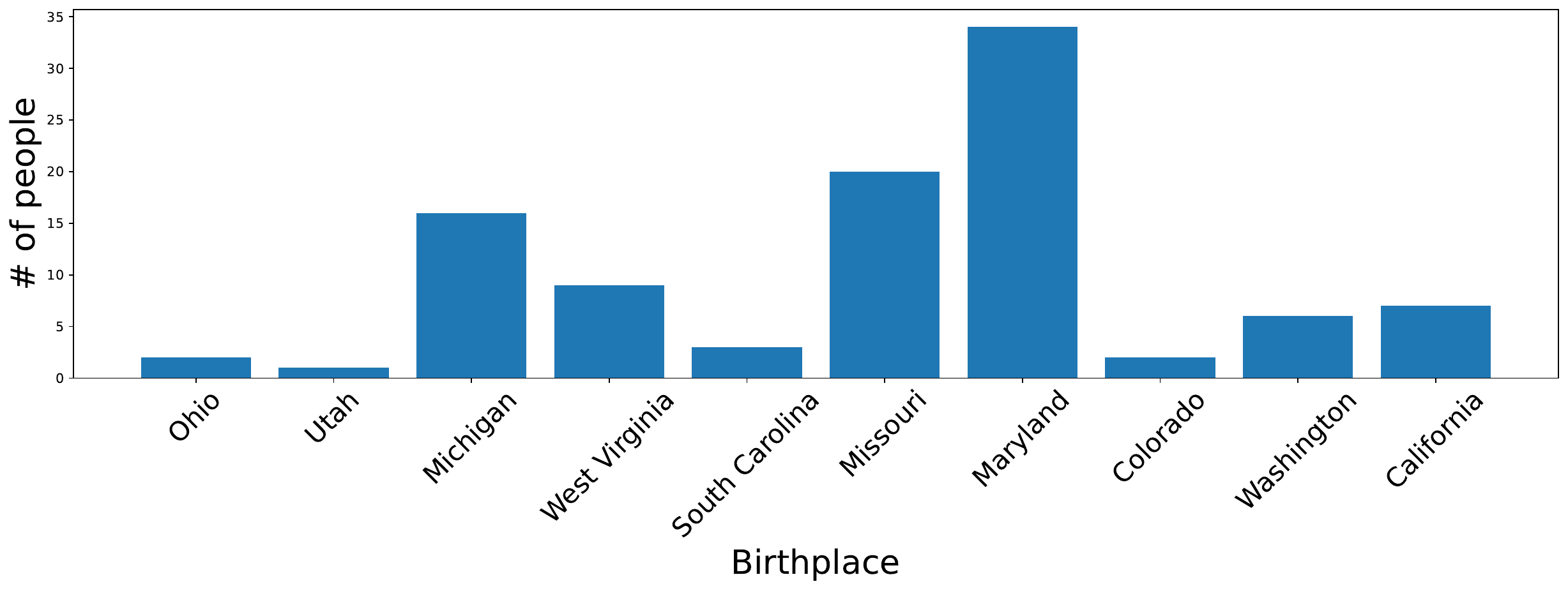}
      \vspace{-4ex}
\caption{Distribution of birthplaces of fictitious people in our synthetic dataset.}
\label{fig:birthplace_dist}

\end{figure}

\paragraph{Statistics of \dataname}
For a better understanding of our synthetic dataset \dataname, we present some statistics here. 
\begin{itemize}
	\item The distribution of family relations Figure~\ref{fig:relation_dist}. It is observed that \textit{child}, \textit{father} and \textit{mother} are top-three relationships in our dataset.
	\item The distribution of the birth year is plotted in Figure~\ref{fig:birthyear_dist}, in a range of 1890 - 2000. 
	\item The set of jobs, collected from the job list across years 1900-2020, such as \{Lawyer, Physician, Sales Manager,..\}.
	\item The distribution of birthplace is summarized in Figure~\ref{fig:birthplace_dist}. 
\end{itemize}

\subsection{Summary of two datasets.} 
The two datasets offer complementary strengths. MQuAKE enables the evaluation of deep unlearning at a realistic setting: over real-world facts using multi-hop reasoning in pretrained models.
\dataname\ instead introduces a fully controlled, setting with realistic logical rules, allowing for reliable evaluation under more complex deductive scenarios.

%% file: sections/experiment.tex
\section{Experiments}
\label{sec:exp}
In this section, we investigate to what extent current unlearning methods succeed at \problemname, and compare the gap between deep unlearning and superficial unlearning.
We release our dataset and code as a benchmark publicly at {\color{blue}\url{https://github.com/wrh14/deep_unlearning}}.
\subsection{Experiment setups}
\paragraph{Unlearning methods.}
We evaluate four common unlearning methods in the literature, similar to the setup in \citet{shi2024muse}; the implementation details such as hyperparameter values are described in Appendix~\ref{sec:app_exp_setting}. 

\textit{Gradient Ascent} (GA; \citep{jang2022knowledge}) maximizes loss on target data, which is a reversed process of learning with gradient \emph{descent}. More optimization steps $T$ result in better unlearning but worse accuracy on extraneous facts. 

\textit{Negative Preference Optimization} (NPO; \citep{zhang2024negative}) optimizes the model $f_{\theta}$ by minimizing the difference between the likelihood of the target data $L(x_{\rm target}; f_{\theta})$ and the likelihood $L(x_{\rm target}; f_{\rm original})$ from the original model $f_{\rm original}$, while not allowing the unlearnt model to diverge too much from the original model. The objective is defined as
$
\mathcal{L}(x_{\rm target}, \theta) = -\frac{2}{\beta}\log \sigma \left(\beta\log\left(\frac{L(x_{\rm target}; f_{\theta})}{L(x_{\rm target}; f_{\rm original})}\right) \right).
$
As suggested by the literature~\citep{rafailov2024direct, zhang2024negative, shi2024muse}, parameter $\beta$ that controls the degree of divergence between unlearnt and original models is set to $0.1$.
Optimization step $T$ is used to control the trade-off between the unlearning and the model utility. 

\textit{Task Vector} (TV; \citep{ilharco2023editing}) first finetunes the original model $f_{\rm original}$ on the target data $x_{\rm target}$ until the original model overfits to the target data. Let $f_{\rm overfit}$ denote the overfitted model. Then the difference $f_{\rm overfit} - f_{\rm original}$ can be used as the direction towards learning $x_{\rm target}$, and its negative direction can be used for unlearning the target data. Therefore, TV defines the unlearning model as $f_{\rm original} - \alpha \cdot \left( f_{\rm overfit} - f_{\rm original}\right)$. A larger value of parameter $\alpha$ gives a higher degree of unlearning but hurts the model utility.

\textit{Who's Harry Potter} (WHP; \citep{eldan2023s}) is based on a similar idea as TV and uses the overfitted model $f_{\rm overfit}$. 
Instead of being guided by the difference in weights it uses the probability. Let $P_{f}(x_t|x_{1:t-1})$ denote the logit vector for predicting the next token $x_t$ from the language model $f$ and prompt $x_{1:t-1}$. WHP samples the next token by the logit vector defined as
\begin{align}
\label{eq:whp}
&P_{f_{\rm original}}(x_t|x_{1:t-1}) - \nonumber\\
& \alpha\cdot \max(P_{f_{\rm overfit}}(x_t|x_{1:t-1}) - P_{f_{\rm original}}(x_t|x_{1:t-1}), 0).
\end{align} 
The role of $\alpha$ is similar to the $\alpha$ in TV.

\paragraph{Target LLMs.} 
For MQuAKE dataset, we test with the \textit{pretrained models} GPT-J-6B and Vicuna-7B where the dataset was built from. 
For Eval-DU, we test with models fine-tuned on the synthetic dataset, which allows us to have more flexible choices of the testing LLMs. We choose four popular LLMs: GPT2-XL~(\citep{radford2019language}, 1.5B) Phi-1.5~(\citep{textbooks2}, 1.3B), Llama2-7b~(\citep{touvron2023llama}, 7B), Llama3-8b~(\citep{dubey2024llama}, 8B). 
We finetune these pre-trained LLMs on our synthetic dataset \dataname; see Appendix~\ref{sec:app_exp_setting} for more finetuning details.
As shown in Table~\ref{tab:ft_utility}, all LLMs have $100\%$ accuracy on the synthetic facts in \dataname, as well as reasonable performance on LLM's general benchmarks, \emph{MMLU}~\citep{hendrycks2021measuring} for multi-domain language understanding, \emph{PIQA}~\citep{bisk2020piqa} for commonsense reasoning, and \emph{RACE}~\citep{lai-etal-2017-race} for reading comprehension.

\begin{table}[!t]
\centering
\caption{Performance of pre-trained models and finetuned models, evaluated with accuracy on the dataset and three LLM benchmarks MMLU, RACE, PIQA.}
\begin{tabular}{c|cccc}
\toprule
	 & Acc. & MMLU & PIQA & RACE \\
	 \midrule
GPT2-XL & 1.0 & 0.23 & 0.71 & 0.33\\
Phi-1.5 & 1.0 & 0.40 & 0.74 & 0.36\\
Llama2-7b & 1.0 & 0.30 & 0.78 & 0.40\\
Llama3-8b & 1.0 & 0.50 & 0.79 & 0.40\\
\midrule
GPT-J-6B & 1.0 & 0.27 & 0.75 & 0.38\\
Vicuna-7B & 1.0 & 0.49 & 0.77 & 0.42\\
\bottomrule
\end{tabular}
      \label{tab:ft_utility}
\end{table}

\paragraph{Target facts for deep unlearning.}
For MQuAKE, we select 82 two-hop and 42 three-hop instances as the knowledge base $\calK$ in interests, that are verifiably present in the pretrained models GPT-J-6B and Vicuna-7B, as identified by the original MQuAKE dataset.
We select 50 such targets from the datasets: 25 from two-hop instances and 25 from three-hop instances. 
We assess each model’s ability to deeply unlearn 50 super-relation facts from these instances.

For Eval-DU, since we have 11 different family relationships (e.g., \textit{child}),  we pick 5 facts for each family relationship, which results in 55 facts in total for the unlearning evaluation.
We will evaluate deep unlearning on unlearning these 55 facts.

\paragraph{Evaluation metrics for the unlearn-model utility trade-off.} The performance of deep unlearning is evaluated by \textit{Success-DU} (Equation~\ref{eq:sucess_du}) and \textit{Recall} (Equation~\ref{eq:recall}), and the model utility is measured by \textit{accuracy} (on our synthetic knowledge base; Equation~\ref{eq:acc}) as well as the utility scores evaluated on three LLM benchmarks MMLU, PIQA and RACE.
For each unlearning method, we can vary its trade-off parameter, and collect a list of success-du, recall, accuracy and utility scores on the three benchmarks. 

To measure the trade-off between unlearning and model utility, we calculate the Success-DU when the accuracy is larger than $0.8$ (Success-DU$@$Acc$\geq 0.8$; $\uparrow$), recall when accuracy is larger than $0.8$ (Recall$@$Acc$\geq 0.8$; $\uparrow$), and Success-DU when the utility score is higher than $95\%$ of the MMLU utility score that the pre-trained / finetuned model (before unlearning) has (Success-DU$@$MMLU$\geq0.95$FT; $\uparrow$). We also report these trade-off evaluations in the appendix for completion: the area under the Accuracy-Recall curve (AR-AUC; $\uparrow$) and  Success-DU$@$U$\geq0.95$FT for the PIQA and RACE utility scores.

\subsection{Main results}
\label{sec:result}

\begin{table*}[!t]
\centering
\caption{Trade-off between deep unlearning and model utility of four unlearning methods on MQuAKE dataset and two pre-trained LLMs. 
}
\vspace{-2ex}
      \label{tab:mquake}
\begin{tabular}{c|cccc|cccc|cccc}
\toprule
Metrics &\multicolumn{4}{c|}{Success-DU@Acc$\geq0.8$ ($\uparrow$)} &\multicolumn{4}{c|}{Recall@Acc$\geq0.8$ ($\uparrow$)} & \multicolumn{4}{c}{Success-DU$@$MMLU$\geq0.95$FT ($\uparrow$)} \\
\midrule
Unlearning methods & GA & NPO & TV & WHP & GA & NPO & TV & WHP & GA & NPO & TV & WHP \\
	 \midrule
 GPT-J-6B & 0.15 & 0.11 & 0.16 & 0.12 & 0.30 & 0.28 & 0.32 & 0.36 & 0.62  & 1.00  & 0.82  & 0.00\\
Vicuna-7B & 0.28 & 0.16 & 0.18 & 0.21 & 0.48 & 0.37 & 0.42 & 0.46 & 0.53  & 1.00  & 0.11  & 0.00\\
\bottomrule
\end{tabular}
\end{table*}

\begin{table*}[!t]
\centering
\caption{Trade-off between deep unlearning and model utility of four unlearning methods on Eval-DU dataset and four fine-tuned LLMs.
}
\vspace{-2ex}
      \label{tab:eval_du}
\begin{tabular}{c|cccc|cccc|cccc}
\toprule
Metrics &\multicolumn{4}{c|}{Success-DU@Acc$\geq0.8$ ($\uparrow$)} &\multicolumn{4}{c|}{Recall@Acc$\geq0.8$ ($\uparrow$)} & \multicolumn{4}{c}{Success-DU$@$MMLU$\geq0.95$FT ($\uparrow$)} \\
\midrule
Unlearning methods & GA & NPO & TV & WHP & GA & NPO & TV & WHP & GA & NPO & TV & WHP \\
	 \midrule
Phi-1.5 & 0.48 & 0.08 & 0.25 & 0.02 & 0.91 & 0.63 & 0.76 & 0.52 & 0.67  & 0.93  & 0.30  & 0.02\\
GPT2-XL & 0.03 & 0.00 & 0.01 & 0.00 & 0.74 & 0.60 & 0.48 & 0.44 & 0.00  & 0.00  & 1.00  & 0.00\\
Llama2-7b & 0.00 & 0.63 & 0.10 & 0.02 & 0.77 & 0.91 & 0.69 & 0.52 & 0.13  & 0.00  & 0.34  & 0.00\\
Llama3-8b & 0.24 & 0.41 & 0.31 & 0.06 & 0.81 & 0.74 & 0.80 & 0.54 & 0.88 & 0.74 & 0.87 & 0.71\\

\bottomrule
\end{tabular}
\end{table*}

\paragraph{Deep unlearning on MQuAKE in Table~\ref{tab:mquake}.} We observe that Success-DU@Acc$\geq$0.8 remains consistently low—ranging from 0.1 to 0.3—for all four unlearning methods across both LLMs. This indicates that even when overall accuracy is maintained, successful deep unlearning is rarely achieved. In contrast, Recall@Acc$\geq$0.8 is notably higher than its corresponding Success-DU values, as recall is a more relaxed, continuous measure that reflects partial unlearning. Lastly, we find that Success-DU@MMLU$\geq$0.95FT is generally higher than Success-DU@Acc$\geq$0.8. In particular, NPO achieves a perfect score of $1$ for both two models, suggesting that preserving general capabilities (e.g., MMLU performance) is substantially easier than preserving unrelated facts within the same knowledge base.

\paragraph{Deep unlearning on Eval-DU in Table~\ref{tab:eval_du}.} On the Eval-DU benchmark, Recall@Acc$\geq$0.8 is generally higher than the value on the MQuAKE benchmark, indicating that many unlearning methods remove a large portion of the minimal deep unlearning set. However, this does not necessarily translate to improved Success-DU@Acc$\geq$0.8. As shown in Appendix Figure~\ref{fig:num_min_dist}, the size of the minimal deep unlearning set ranges from 1 to 17. In many cases, a method may successfully unlearn most \emph{but not all} supporting facts, resulting in high recall but zero Success-DU. This explains why some methods perform poorly in terms of Success-DU, with values near zero on certain LLMs.

\paragraph{Summary.}
Deep unlearning on MQuAKE represents a relatively simple setting, as it requires removing only one additional fact per target (i.e., a minimal set size of 2). Yet, even in this mild regime, Success-DU scores remain low, highlighting the challenge of achieving complete unlearning. Eval-DU presents a more complex scenario, where the minimal unlearning sets can exceed 10 facts, further increasing the difficulty. 
Together, these results show that existing unlearning methods struggle to guarantee full removal of logically implied knowledge: \textbf{no unlearning method achieves Success-DU above $0.8$ and at the same time keeps the retain accuracy above $0.8$.}

\begin{table*}[!t]
\centering
\caption{Success-SU@Acc$\geq 0.8$ ($\uparrow$) (the success rate of \textbf{superficial unlearning} at $80\%$ accuracy), where only the target fact itself is required to be unlearnt. We also present $\Delta$ = (Success-SU@Acc$\geq 0.8$ - Success-DU@Acc$\geq 0.8$) in color {\color{blue}blue}.
}
\label{tab:sup_numbers}
\begin{tabular}{c|c|cccc}
\toprule
	\multicolumn{2}{c|}{Unlearning methods} & GA & NPO & TV & WHP \\
	\midrule
\multirow{2}{*}{MQuAKE} & GPT-J-6B & 0.32 ({\color{blue}+0.17}) & 0.21 ({\color{blue}+0.10}) & 0.35 ({\color{blue}+0.19}) & 0.33 ({\color{blue}+0.21})\\
 & Vicuna-7B & 0.58 ({\color{blue}+0.30}) & 0.44 ({\color{blue}+0.28}) & 0.61 ({\color{blue}+0.43}) & 0.53 ({\color{blue}+0.32})\\
	 \midrule
\multirow{4}{*}{Eval-DU} & Phi-1.5 & 0.59 ({\color{blue}+0.11}) & 0.46 ({\color{blue}+0.38}) & 0.84 ({\color{blue}+0.59}) & 0.55 ({\color{blue}+0.53})\\
 & GPT2-XL & 0.80 ({\color{blue}+0.77}) & 0.59 ({\color{blue}+0.59}) & 0.60 ({\color{blue}+0.59}) & 0.59 ({\color{blue}+0.59})\\
 & Llama2-7b & 0.81 ({\color{blue}+0.81}) & 0.85 ({\color{blue}+0.22}) & 0.83 ({\color{blue}+0.73}) & 0.44 ({\color{blue}+0.42})\\
 & Llama3-8b & 0.79 ({\color{blue}+0.55}) & 0.75 ({\color{blue}+0.34}) & 0.79 ({\color{blue}+0.48}) & 0.47 ({\color{blue}+0.41})\\
\bottomrule
\end{tabular}
\end{table*}
 
\begin{table*}[!t]
\centering
\caption{Success-DU@Acc$\geq0.8$ of four unlearning methods on two benchmarks when assuming the knolwedge of minimal deep unlearning sets for each unlearning target. We also present $\Delta$ = (Success-SU@Acc$\geq 0.8$ - Success-DU@Acc$\geq 0.8$) in color {\color{blue}blue} if $\Delta\geq 0$ or color {\color{red}red} if $\Delta< 0$.
}
      \label{tab:white_box}
\begin{tabular}{c|c|cccc}
\toprule
\multicolumn{2}{c|}{Unlearning methods} & GA & NPO & TV & WHP \\
\midrule
\multirow{2}{*}{MQuAKE} & GPT-J-6B & 0.11 ({\color{red}-0.04}) & 0.13 ({\color{blue}+0.02}) & 0.17 ({\color{blue}+0.01}) & 0.12 ({\color{blue}+0.00})\\
 & Vicuna-7B & 0.31 ({\color{blue}+0.03}) & 0.19 ({\color{blue}+0.03}) & 0.20 ({\color{blue}+0.18}) & 0.26 ({\color{blue}+0.05})\\
	 \midrule
\multirow{4}{*}{Eval-DU}& Phi-1.5 & 0.46 ({\color{red}-0.03}) & 0.11 ({\color{blue}+0.03}) & 0.27 ({\color{blue}+0.02}) & 0.18 ({\color{blue}+0.16}) \\
& GPT2-XL & 0.13 ({\color{blue}+0.10}) & 0.20 ({\color{blue}+0.20}) & 0.10 ({\color{blue}+0.09}) & 0.10 ({\color{blue}+0.10})\\
& Llama2-7b & 0.09 ({\color{blue}+0.09}) & 0.63 ({\color{blue}+0.00}) & 0.25 ({\color{blue}+0.15}) & 0.10 ({\color{blue}+0.08})\\
& Llama3-8b & 0.43 ({\color{blue}+0.19}) & 0.48 ({\color{blue}+0.07}) & 0.25 ({\color{red}-0.06}) & 0.17 ({\color{blue}+0.11})\\

\bottomrule
\end{tabular}
\end{table*}

\subsection{Superficial unlearning versus \problemname.}
We examine the distinction between superficial unlearning, where the target fact is no longer directly retained, and deep unlearning, which requires removing all deducible paths to the target fact. Specifically, we measure the success rate of superficial unlearning when the post-unlearning model retains at least $0.8$ accuracy on unrelated facts (Success-SU@Acc$\geq$0.8).

As shown in Table~\ref{tab:sup_numbers}, Success-SU is consistently much higher than Success-DU in Table~\ref{tab:eval_du}, demonstrating that \textbf{deep unlearning is substantially more demanding than superficial unlearning}. For example, when applying GA to GPT2-XL on the Eval-DU benchmark, Success-SU reaches 0.80, whereas the corresponding Success-DU is only 0.03. 
Moreover, \textbf{this gap is generally larger on the Eval-DU benchmark than on MQuAKE}.
The gap on Eval-DU can be around $0.8$, while the gap (presented in blue in Table~\ref{tab:sup_numbers}) on MQuAKE is at most $0.3$.
We hypothesize that this can be attributed to the larger sizes of minimal deep unlearning sets in Eval-DU.

\section{White-box Deep Unlearning}
\label{sec:white_box}
In this section, we explore a white-box variant of the deep unlearning problem, where the unlearning algorithm is given access to the minimal deep unlearning set for each target fact. If significant improvements are observed, this would motivate future research on reliably identifying such minimal sets prior to unlearning.
We adopt the same experimental setup as in Section~\ref{sec:exp}, but modify the inputs to each unlearning method: instead of unlearning only the target facts, we jointly unlearn any entire minimal deep unlearning set for each target. Table~\ref{tab:white_box} reports the resulting Success-DU@Acc$\geq$0.8 scores across both benchmarks.

Compared to the original results in Table~\ref{tab:mquake} and Table~\ref{tab:eval_du}, we observe consistent improvements across all methods and settings. 
Notably, the gains are most pronounced for configurations that initially had relatively low Success-DU scores.
In contrast, configurations that already performed well (e.g., NPO on Llama2-7b) show little improvement.
This suggests an upper bound in effectiveness even with perfect knowledge of the unlearning set.
We hypothesize such the existence of such upper bound is due to the inherent hardness of large-batch unlearning -- deep unlearning requires to unlearn much more facts than only the target facts.

These results underscore the critical role of identifying minimal unlearning sets and motivate future work on the automated discovery of such deductive structures. 
Furthermore, they point to a second future direction: developing algorithms that go beyond simply taking the minimal sets as input to overcome current performance limitations.

%% file: sections/related_work.tex
\section{Related Work}
\label{sec:related_work}
\textbf{Benchmarks and evaluations in LLM unlearning.} 
TOFU~\citep{tofu2024} is a benchmark containing fictitious authors and their related biographic question-answering texts, and evaluates the unlearning by comparing the answer from LLM given the question and the ground truth.
WMDP~\citep{li2024the} provides knowledge in biosecurity, cybersecurity, and chemical security, which matches the realistic desire for studying unlearning.
A more recent benchmark MUSE~\citep{shi2024muse} in the domain of news articles and books enriches the evaluation by introducing metrics from both memorization and privacy leakage aspects.
\citet{yao2024machine} introduces a benchmark of evaluating the unlearning in pre-trained data and the metric of unlearning utility is to compute the perplexity of the data from the memorization aspect.
\citet{patil2024can} and \citet{lucki2024adversarial} evaluate the unlearning from an adversarial attack aspect of knowledge extraction.
\citet{joshi2024towards} creats multiple QAs in different formats for checking if the unlearning target is still retained.
\citet{lynch2024eight} introduces eight different methods to evaluate unlearning robustly in llms.
This branch of work focuses on proposing more realistic domains and more robust ways to evaluate the unlearning, and the challenge at their benchmark is to unlearn a large batch of facts or texts while keeping the model utility.
However, none of them consider the interrelation between the target facts and other facts also in the LLM, which our paper focuses on.

\textbf{Unlearning methods in LLM.} In addition to the methods evaluated in Section~\ref{sec:exp}, one popular extension is assuming the existence of a ``retain" set independent of the target facts. When doing gradient ascent or other gradient-based variants, \citet{yao2023large} and \citet{chen2023unlearn} minimize the loss on the ``retain" set simultaneously to avoid quickly losing other irrelevant facts and hence help with the model utility.
Another category is the model-editing based~\citep{de2021editing, meng2022locating, meng2023massediting, wang2024large}, which hypothesizes that the knowledge is saved in certain MLPs in the transformer and proposes an explicit-form solution for the weight update to unlearn the target facts. 
A recent paradigm is in-context unlearning~\citep{pawelczyk2024incontext}, which provides specific kinds of inputs in context rather than editing the model.

%
%

%% file: sections/discussion.tex
\section{Conclusion and Future Work}
\paragraph{Conclusion.} In this paper, we propose a new setting for machine unlearning, referred to as \textit{\problemname}, which emphasizes not only removing a target fact but also eliminating its deductive inferability from retained knowledge. 
To support this setting, we benchmark deep unlearning using both a real-world dataset (MQuAKE) and a newly constructed semi-synthetic dataset (\dataname). 
Our empirical results show that existing unlearning methods often fail to achieve deep unlearning, likely due to overlooking deductive dependencies among facts.

\paragraph{Future work.} This work opens several promising directions for future research.
Firstly, more effective methods can be developed for the \problemname\ setting, with an awareness of connections between facts.
In particular, Section~\ref{sec:white_box} highlights the value of identifying minimal unlearning sets, suggesting the potential of automating the discovery of such deductive structures.
Additionally, there is potential to extend our framework with more expressive knowledge representations and sophisticated deductive processes beyond the current scope of triplet facts and logical rules.

%% file: sections/appendix.tex
\subsection{More Details on Experimental Settings and More Experimental Results}
\label{sec:app_exp_setting}
\paragraph{Details of finetuning LLMs on \dataname.} 
The finetuning is under the question-answering format, where the question is given in the prompt and the loss is computed from the answer. 
The batch size of finetuning on all four LLMs is 16.
The learning rate is $2e-5$ for GPT-XL and Phi-1.5 and $1e-5$ for Llama2-7b and Llama3-8b; the learning rate scheduler is the linear scheduler from HuggingFace~\citep{wolf2019huggingface}.
The number of epochs is $10$ for Phi-1.5, Llama2-7b, and Llama3-8b and $15$ GPT-XL to guarantee a full memorization after finetuning. 

\paragraph{Details of hyperparameters in unlearning methods.}
For each method, we pick the values of hyperparameter for best reflecting the trade-off.
For GA, the learning rate is $2e-5$ for GPT-XL and Phi-1.5 and $1e-5$ for Llama2-7b and Llama3-8b; the learning rate scheduler is the linear scheduler from HuggingFace~\citep{wolf2019huggingface}.
The hyperparameter of the optimization iteration $T$ is selected from $\{1, 2, 4, 8, 16\}$ for Phi-1.5, Llama2-7b and Llama3-8b and $\{1, 2, 4, 8, 16, 32\}$ for GPT-XL.
For NPO, the learning rate is $4e-5$ for GPT-XL and Phi-1.5 and $2e-5$ for Llama2-7b and Llama3-8b; the learning rate scheduler is the linear scheduler from HuggingFace~\citep{wolf2019huggingface}.
The hyperparameter of the optimization iteration $T$ is selected from $\{1, 2, 4, 8, 16\}$ for Phi-1.5, Llama2-7b and Llama3-8b and $\{1, 2, 4, 8, 16, 32\}$ for GPT-XL.
For both TV and WHP, the ``overfit" model is finetuned with 10 more iterations on the target data point.
In TV, the hyperparameter $\alpha$ is from $\{0.2, 1.0, 5.0, 10.0, 30.0, 60.0, 80.0\}$ for GPT-XL and $\{0.2, 0.5, 1.0, 5.0, 10.0\}$ for Phi-1.5, Llama2-7b and Llama3-8b.
In WHP, the hyperparameter $\alpha$ is from $\{0.5, 1.0, 5.0, 10.0, 100.0, 1000.0\}$.

\begin{figure*}[!t]
\centering
\begin{minipage}{.23\linewidth}
      \centering
      \includegraphics[width=\linewidth]{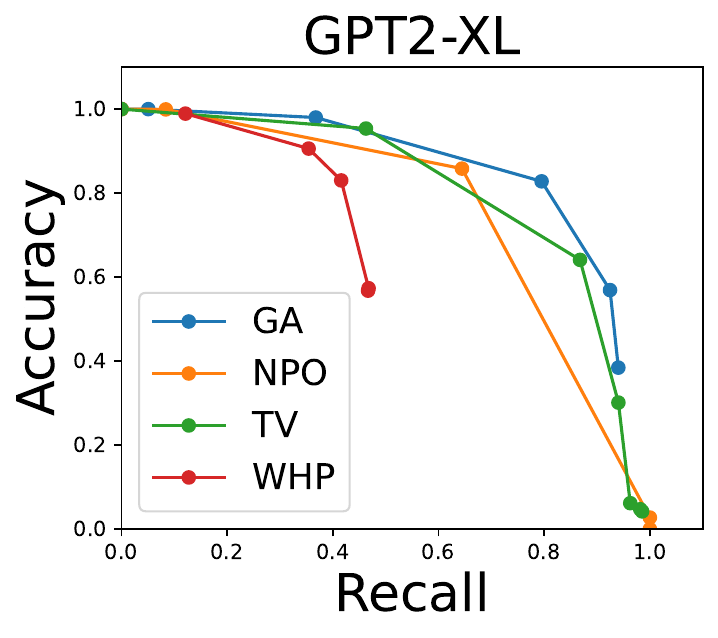}
\end{minipage}
\begin{minipage}{.23\linewidth}
      \centering
      \includegraphics[width=\linewidth]{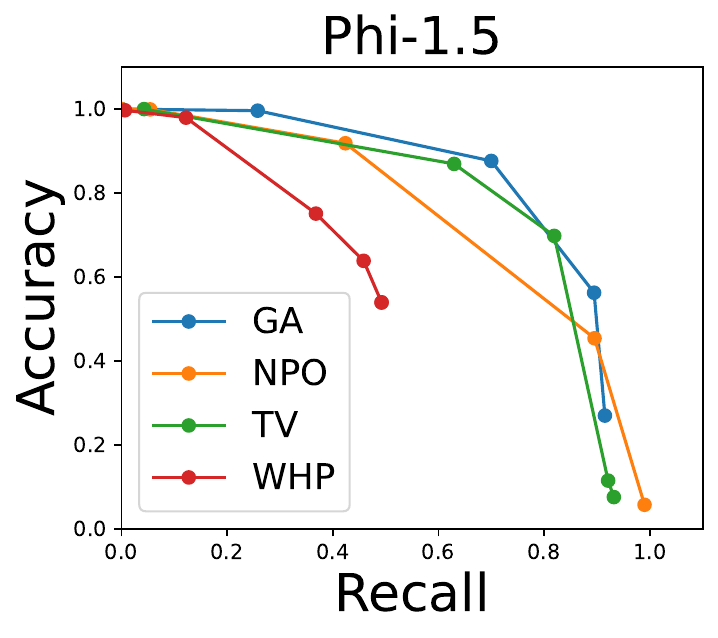}
\end{minipage}
\begin{minipage}{.23\linewidth}
      \centering
      \includegraphics[width=\linewidth]{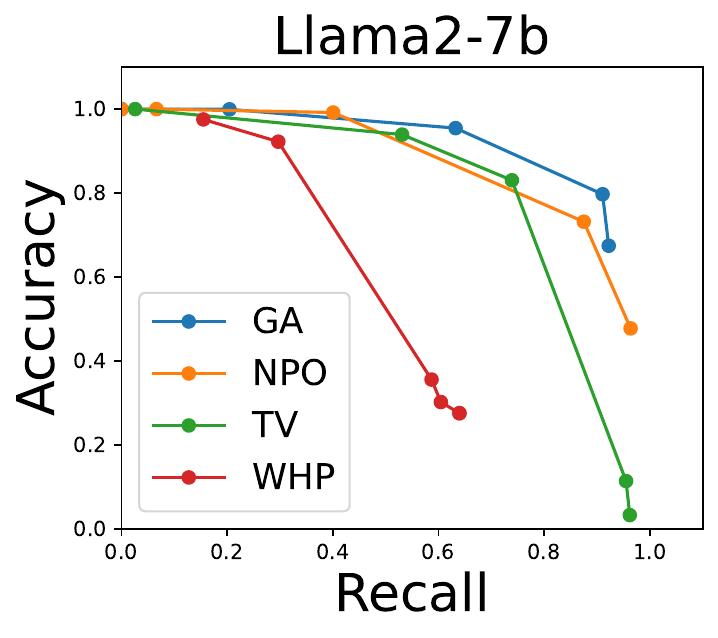}
\end{minipage}
\begin{minipage}{.23\linewidth}
      \centering
      \includegraphics[width=\linewidth]{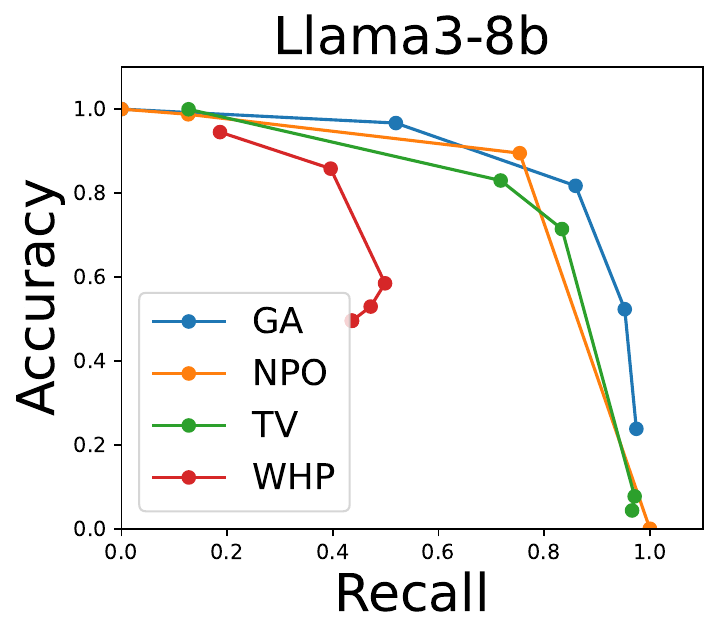}
\end{minipage}
\caption{Accuracy-Recall curve when testing four methods for \problemnameing\ from four LLMs.}
\vspace{-1ex}
\label{fig:curve_acc_recall_all}
\end{figure*}

\begin{figure*}[!t]
\centering
\begin{minipage}{.23\linewidth}
      \centering
      \includegraphics[width=\linewidth]{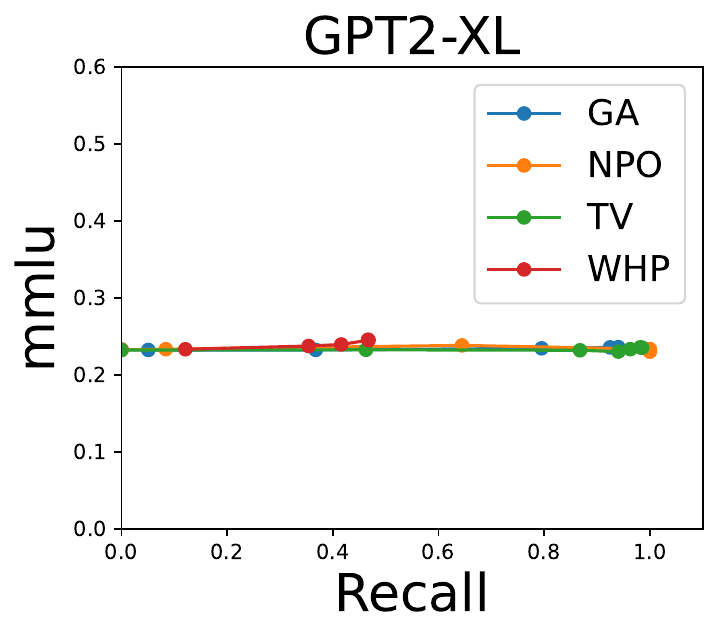}
\end{minipage}
\begin{minipage}{.23\linewidth}
      \centering
      \includegraphics[width=\linewidth]{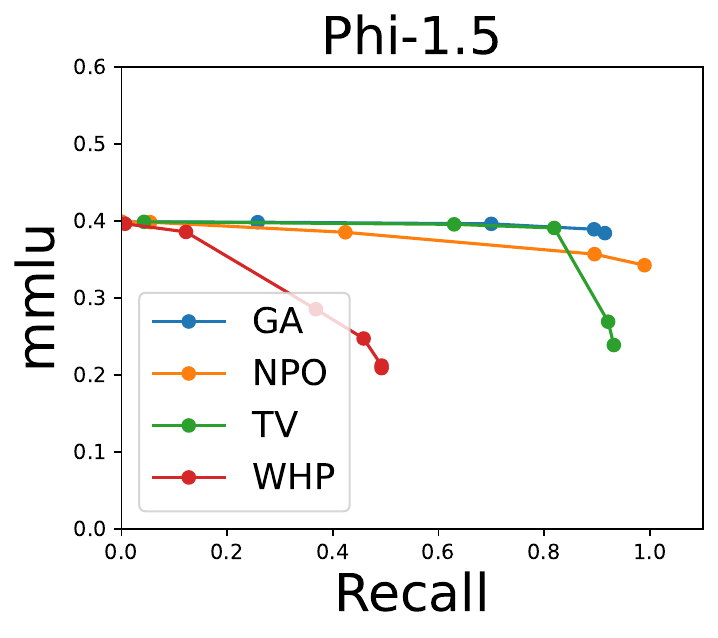}
\end{minipage}
\begin{minipage}{.23\linewidth}
      \centering
      \includegraphics[width=\linewidth]{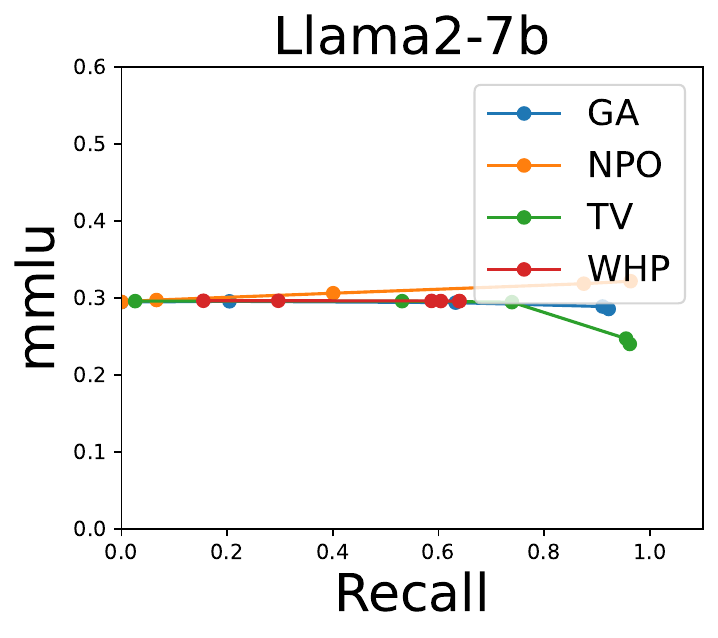}
\end{minipage}
\begin{minipage}{.23\linewidth}
      \centering
      \includegraphics[width=\linewidth]{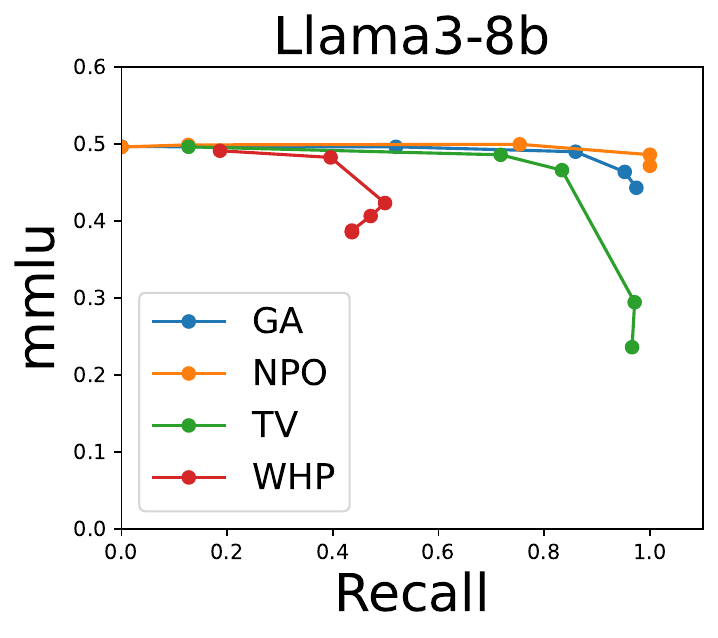}
\end{minipage}
\caption{MMLU-Recall curve when testing four methods for \problemnameing\ from four LLMs.}
\vspace{-1ex}
\label{fig:curve_mmlu_recall_all}
\end{figure*}

\begin{figure*}[!t]
\centering
\begin{minipage}{.23\linewidth}
      \centering
      \includegraphics[width=\linewidth]{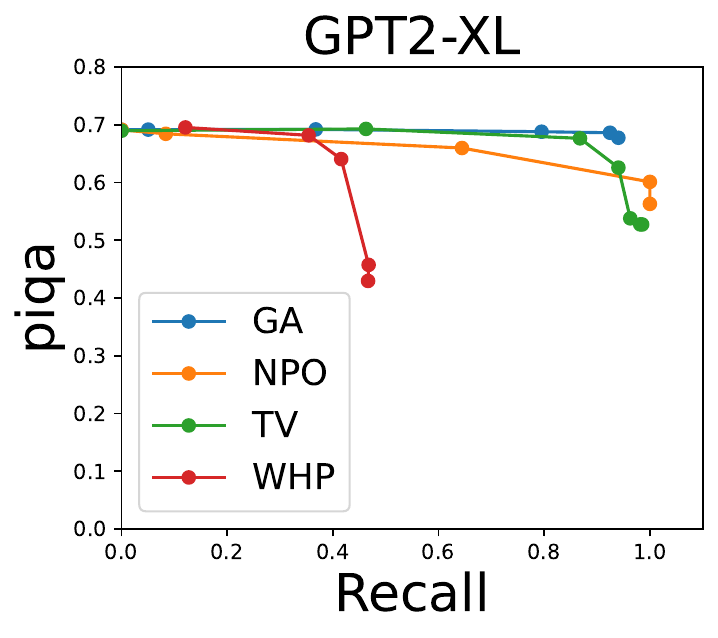}
\end{minipage}
\begin{minipage}{.23\linewidth}
      \centering
      \includegraphics[width=\linewidth]{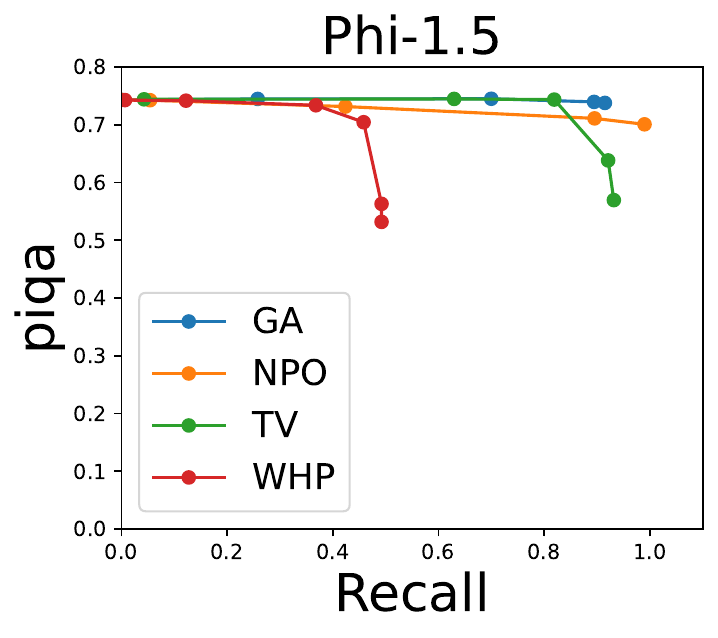}
\end{minipage}
\begin{minipage}{.23\linewidth}
      \centering
      \includegraphics[width=\linewidth]{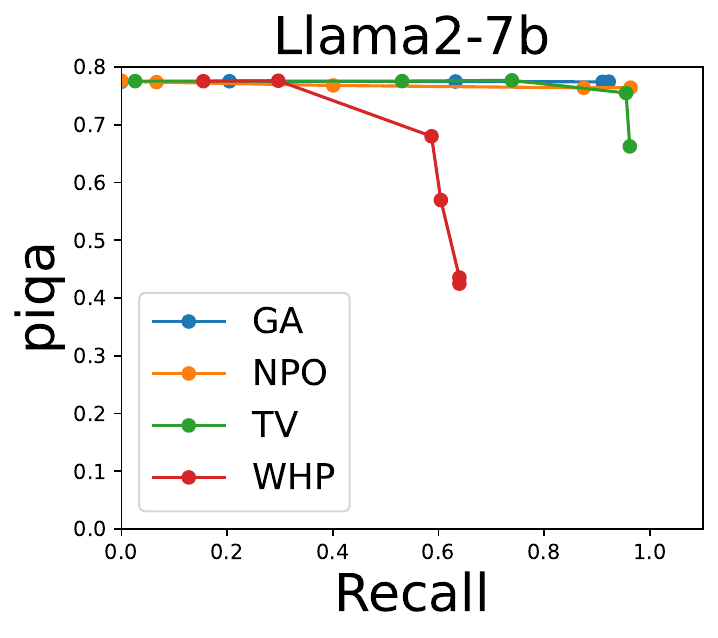}
\end{minipage}
\begin{minipage}{.23\linewidth}
      \centering
      \includegraphics[width=\linewidth]{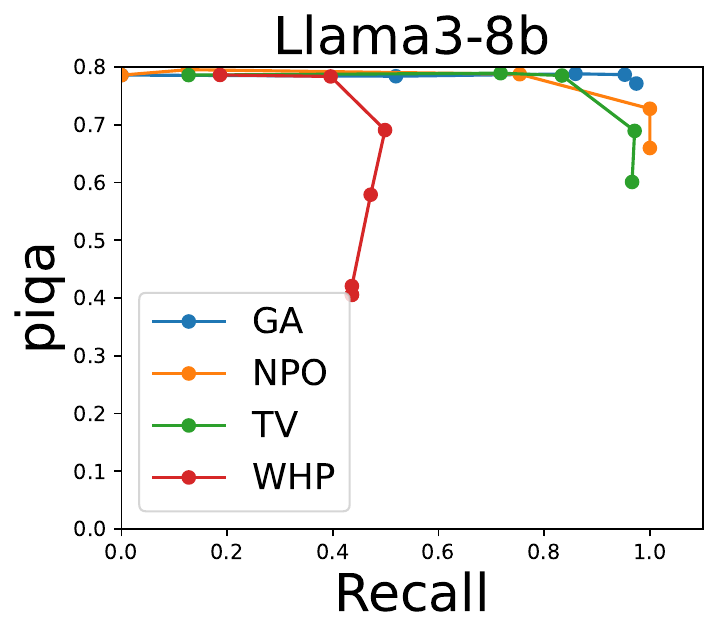}
\end{minipage}
\caption{PIQA-Recall curve when testing four methods for \problemnameing\ from four LLMs.}
\vspace{-1ex}
\label{fig:curve_piqa_recall_all}
\end{figure*}

\begin{figure*}[!t]
\centering
\begin{minipage}{.23\linewidth}
      \centering
      \includegraphics[width=\linewidth]{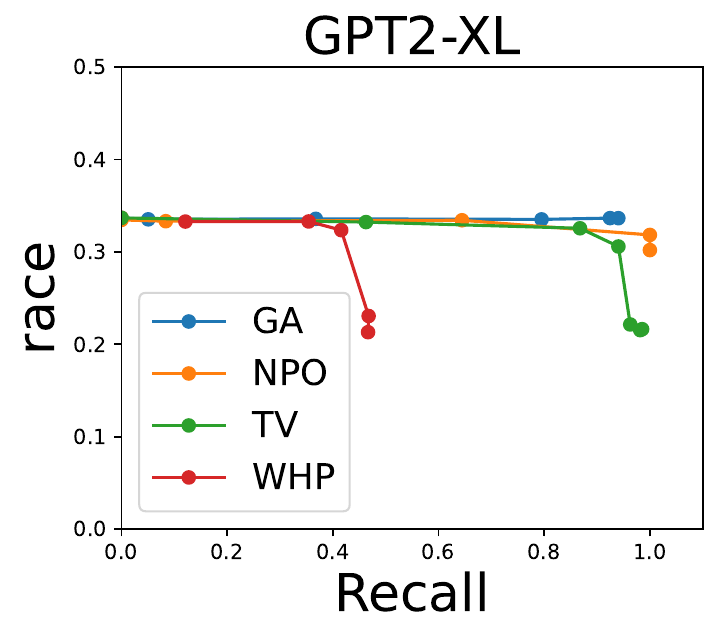}
\end{minipage}
\begin{minipage}{.23\linewidth}
      \centering
      \includegraphics[width=\linewidth]{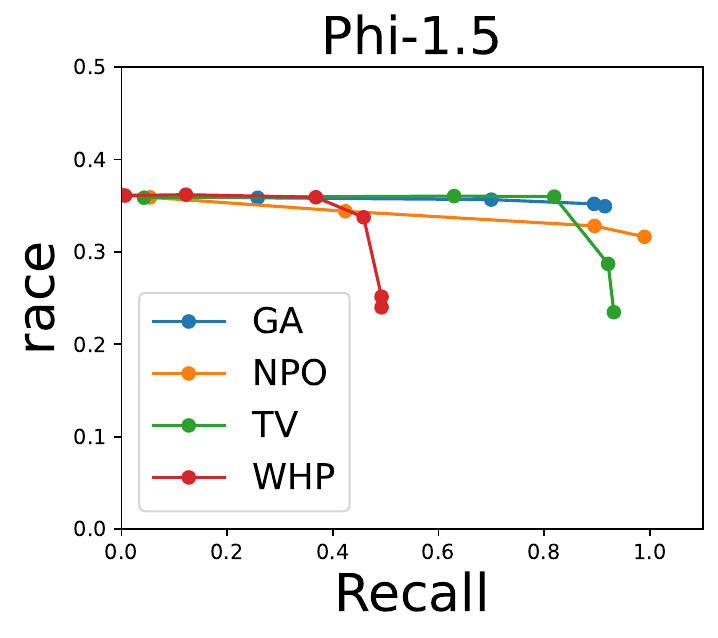}
\end{minipage}
\begin{minipage}{.23\linewidth}
      \centering
      \includegraphics[width=\linewidth]{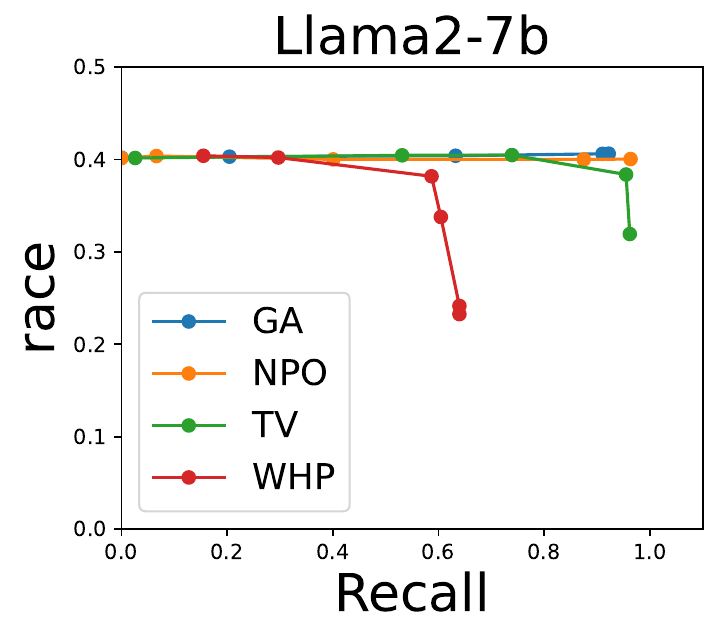}
\end{minipage}
\begin{minipage}{.23\linewidth}
      \centering
      \includegraphics[width=\linewidth]{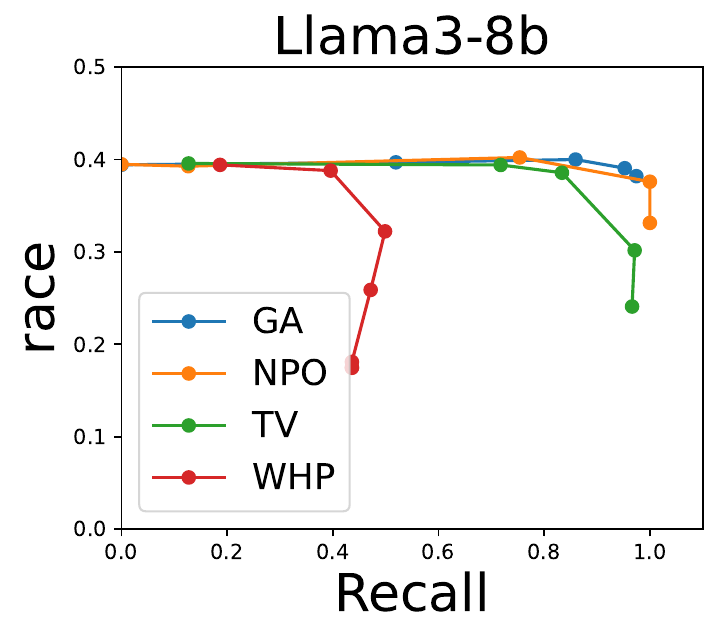}
\end{minipage}
\caption{RACE-Recall curve when testing four methods for \problemnameing\ from four LLMs.}
\vspace{-1ex}
\label{fig:curve_race_recall_all}
\end{figure*}

\paragraph{Trade-off curves of four unlearning methods on four LLMs.}
In the main paper, we have presented Accuracy-Recall curve and MMLU-Recall curve of four unlearning methods on Phi-1.5. In this section, we show the Accuracy-Recall curves on all four LLMs in Figure~\ref{fig:curve_acc_recall_all} and the trade-off curve between utility scores on three benchmarks (MMLU, PIQA, RACE) and Recall in Figure~\ref{fig:curve_mmlu_recall_all}, Figure~\ref{fig:curve_piqa_recall_all} and Figure~\ref{fig:curve_race_recall_all} respectively.

\paragraph{More trade-off evaluation.} Table~\ref{tab:numbers} and Table~\ref{tab:more_utility_du} show the full unlearn-retain trade-off evaluations.

\begin{table*}[!t]
\centering
\caption{Trade-off between deep unlearning and accuracy of four unlearning methods on four LLMs. Particularly to evaluate the trade-off, we measure Success-DU@ACC$\geq0.8$, Recall@Acc$\geq0.8$ and AR-AUC. For each metric and LLM, we highlight the best score achieved by any unleanring method.
}
      \label{tab:numbers}
\begin{tabular}{c|c|cccc|cccc|cccc}
\toprule
\multicolumn{2}{c|}{Metrics} &\multicolumn{4}{c|}{Success-DU@ACC$\geq0.8$ ($\uparrow$)} &\multicolumn{4}{c|}{Recall@Acc$\geq0.8$ ($\uparrow$)} & \multicolumn{4}{c}{AR-AUC ($\uparrow$)} \\
\midrule
\multicolumn{2}{c|}{Unlearning methods} & GA & NPO & TV & WHP & GA & NPO & TV & WHP & GA & NPO & TV & WHP \\
	 \midrule
\multirow{4}{*}{Eval-DU} & Phi-1.5 & 0.48 & 0.08 & 0.25 & 0.02 & 0.91 & 0.63 & 0.76 & 0.52 & 0.92 & 0.81 & 0.82 & 0.69\\
 & GPT2-XL & 0.03 & 0.00 & 0.01 & 0.00 & 0.74 & 0.60 & 0.48 & 0.44 & 0.78 & 0.80 & 0.62 & 0.59\\
 & Llama2-7b & 0.00 & 0.63 & 0.10 & 0.02 & 0.77 & 0.91 & 0.69 & 0.52 & 0.87 & 0.92 & 0.77 & 0.63\\
 & Llama3-8b & 0.24 & 0.41 & 0.31 & 0.06 & 0.81 & 0.74 & 0.80 & 0.54 & 0.88 & 0.74 & 0.87 & 0.71\\
\midrule
\multirow{2}{*}{MQuAKE} & GPT-J-6B & 0.15 & 0.11 & 0.16 & 0.12 & 0.30 & 0.28 & 0.32 & 0.36 & 0.62 & 0.56 & 0.63 & 0.67\\
 & Vicuna-7B & 0.28 & 0.16 & 0.18 & 0.21 & 0.48 & 0.37 & 0.42 & 0.46 & 0.70 & 0.61 & 0.71 & 0.68\\
\bottomrule
\end{tabular}
\end{table*}

\begin{table*}[!t]
\caption{Trade-off between deep unlearning and utility scores on three benchmarks MMLU, PIQA, and RACE. The metric for evaluating the trade-off is Success-DU$@$U$\geq0.95$FT ($\uparrow$). For each benchmark and each LLM, we highlight the best score achieved by any unleanring method.
}
\centering
      \label{tab:more_utility_du}
\begin{tabular}{c|c|cccc|cccc|cccc}
\toprule
\multicolumn{2}{c|}{LLM benchmarks} &\multicolumn{4}{c|}{MMLU} & \multicolumn{4}{c|}{PIQA} & \multicolumn{4}{c}{RACE} \\
\midrule
\multicolumn{2}{c|}{Unlearning methods} & GA & NPO & TV & WHP & GA & NPO & TV & WHP & GA & NPO & TV & WHP \\
\midrule
\multirow{4}{*}{Eval-DU} & Phi-1.5  & 0.67  & 0.93  & 0.30  & 0.02  & 0.67  & 0.73  & 0.40  & 0.01  & 0.67  & 0.85  & 0.28  & 0.00\\
 & GPT2-XL  & 0.00  & 0.00  & 1.00  & 0.00  & 0.04  & 0.00  & 0.07  & 0.00  & 0.00  & 0.00  & 0.23  & 0.00\\
 & Llama2-7b  & 0.13  & 0.00  & 0.34  & 0.00  & 0.00  & 1.00  & 0.41  & 0.02  & 0.00  & 1.00  & 0.32  & 0.02\\
 & Llama3-8b  & 1.00  & 0.00  & 0.38  & 0.00  & 1.00  & 1.00  & 0.42  & 0.07  & 0.76  & 0.00  & 0.40  & 0.08\\
\midrule
\multirow{2}{*}{MQuAKE} & GPT-J-6B  & 0.62  & 1.00  & 0.82  & 0.00  & 0.73  & 0.78  & 0.86  & 0.66  & 0.86  & 1.00  & 0.84  & 0.66\\
 & Vicuna-7B  & 0.53  & 1.00  & 0.11  & 0.00  & 0.30  & 1.00  & 0.89  & 0.42  & 0.52  & 0.98  & 0.89  & 0.32\\\bottomrule
\end{tabular}
\vspace{-1ex}
\end{table*}

\begin{table*}[!t]
\centering
\caption{Trade-off between deep unlearning and accuracy of four unlearning methods on four LLMs. Particularly to evaluate the trade-off, we measure Success-DU@ACC$\geq0.8$, Recall@Acc$\geq0.8$ and AR-AUC. For each metric and LLM, we highlight the best score achieved by any unleanring method.
}
      \label{tab:white_box}
\resizebox{\linewidth}{!}{%
\begin{tabular}{c|c|cccc|c|cccc|c|cccc|c}
\toprule
\multicolumn{2}{c|}{Metrics} &\multicolumn{5}{c|}{Success-DU@ACC$\geq0.8$ ($\uparrow$)} &\multicolumn{5}{c|}{Recall@Acc$\geq0.8$ ($\uparrow$)} & \multicolumn{5}{c}{AR-AUC ($\uparrow$)} \\
\midrule
\multicolumn{2}{c|}{Unlearning methods} & GA & NPO & TV & WHP & GD & GA & NPO & TV & WHP & GD & GA & NPO & TV & WHP & GD \\
	 \midrule
\multirow{2}{*}{MQuAKE} & GPT-J-6B & 0.11 & 0.13 & 0.17 & 0.12 & 0.15 & 0.26 & 0.25 & 0.29 & 0.35 & 0.32 & 0.61 & 0.55 & 0.64 & 0.57 & 0.63\\
 & Vicuna-7B & 0.31 & 0.19 & 0.20 & 0.26 & 0.18 & 0.46 & 0.35 & 0.48 & 0.50 & 0.43 & 0.71 & 0.59 & 0.73 & 0.72 & 0.71\\
\midrule
\multirow{4}{*}{Eval-DU}& Phi-1.5 & 0.46 & 0.11 & 0.27 & 0.18 & 0.72 & 0.86 & 0.62 & 0.70 & 0.18 & 0.94 & 0.87 & 0.82 & 0.78 & 0.49 & 0.94\\
& GPT2-XL & 0.13 & 0.20 & 0.10 & 0.10 & 0.25 & 0.80 & 0.59 & 0.64 & 0.64 & 0.83 & 0.82 & 0.80 & 0.70 & 0.78 & 0.89\\
& Llama2-7b & 0.09 & 0.63 & 0.25 & 0.10 & 0.55 & 0.39 & 0.65 & 0.77 & 0.56 & 0.69 & 0.71 & 0.85 & 0.83 & 0.70 & 0.81\\
& Llama3-8b & 0.43 & 0.48 & 0.25 & 0.17 & 0.62 & 0.81 & 0.66 & 0.68 & 0.46 & 0.91 & 0.91 & 0.86 & 0.85 & 0.66 & 0.93\\

\bottomrule
\end{tabular}
}
\vspace{-2ex}
\end{table*}
%